\documentclass[11pt]{article}

\usepackage[final]{acl}

\usepackage{times}
\usepackage{latexsym}

\usepackage[T1]{fontenc}
\newcommand{\myline}[1]{{\medskip\noindent\textbf{#1}}}
\usepackage{multicol}

\usepackage{xspace}
\usepackage{booktabs}
\usepackage{pifont}

\newcommand{\system}{\textsc{IGenBench}\xspace}

\usepackage{multirow}
\usepackage{makecell}
\usepackage[utf8]{inputenc}

\usepackage{microtype}
\usepackage[ruled,lined,linesnumbered]{algorithm2e}
\usepackage{amsmath}
\usepackage{amssymb}
\usepackage{inconsolata}

\usepackage{graphicx}
\usepackage{booktabs}  % 画三线表必备
\usepackage{tabularx}  % 自动换行表格必备
\usepackage{ragged2e}  % 优化表格内的对齐

\usepackage{multirow}
\usepackage{xcolor}   % 支持颜色
\usepackage{colortbl} % 支持单元格着色
\usepackage{arydshln} % 支持虚线 \hdashline
\usepackage{fontawesome5} % 引入图标 (Math, Chart, etc.)
\usepackage{graphicx}

\definecolor{bestblue}{RGB}{220, 235, 250}  % 第一名：淡蓝色
\definecolor{secondgreen}{RGB}{235, 245, 230} % 第二名：淡绿色

\usepackage{enumitem}

\title{\sc{IGenBench: Benchmarking the Reliability of Text-to-Infographic Generation}}

\author{
\normalfont
Yinghao Tang\textsuperscript{1}, Xueding Liu\textsuperscript{2}, Boyuan Zhang\textsuperscript{1}, Tingfeng Lan\textsuperscript{3}, Yupeng Xie\textsuperscript{4}, \\
Jiale Lao\textsuperscript{5}, Yiyao Wang\textsuperscript{1}, Haoxuan Li\textsuperscript{1}, Tingting Gao\textsuperscript{6}, Bo Pan\textsuperscript{1}, 
Luoxuan Weng\textsuperscript{1}, \\ Xiuqi Huang\textsuperscript{1}\thanks{Xiuqi Huang, Yingchaojie Feng, and Yuyu Luo are corresponding authors.}, Minfeng Zhu\textsuperscript{6}, Yingchaojie Feng\textsuperscript{7}\footnotemark[1], 
Yuyu Luo\textsuperscript{4}\footnotemark[1], Wei Chen\textsuperscript{1}
\\
\normalsize
\textsuperscript{1}\textit{State Key Lab of CAD\&CG, Zhejiang University}, 
\\
\textsuperscript{2}\textit{UESTC},
\textsuperscript{3}\textit{University of Virginia},
\textsuperscript{4}\textit{HKUST(GZ)},
\\
\textsuperscript{5}\textit{Cornell University},
\textsuperscript{6}\textit{Zhejiang University},
\textsuperscript{7}\textit{National University of Singapore}
\normalsize
\\
\normalsize
}

\begin{document}
\maketitle
%\begin{abstract}
 %   Infographics enhance data visualizations by integrating illustrative elements, making them indispensable across journalism, education, and business analytics. While recent text-to-image (T2I) models can generate aesthetically expressive images, it remains unclear whether they can reliably generate \emph{correct} infographics that precisely adhere to semantic requirements and accurately encode data values. We present \system, the first comprehensive benchmark for evaluating infographic generation fidelity, comprising 734 curated prompt-infographic pairs covering 30 real-world infographic types. We establish a taxonomy of 10 question types covering all critical infographic elements and decompose each prompt into atomic yes/no verification questions. Our evaluation framework yields question-level accuracy (Q-ACC), measuring the correctness of individual constraints, and infographic-level accuracy (P-ACC), measuring whether all constraints are simultaneously satisfied. Experiments on 10 state-of-the-art T2I models demonstrate that current T2I models cannot yet reliably generate correct infographics.  We release \system at \url{https://igen-bench.vercel.app/}.
%\end{abstract}

\begin{abstract}
Infographics are composite visual artifacts that combine data visualizations with textual and illustrative elements to communicate information. While recent text-to-image (T2I) models can generate aesthetically appealing images, their reliability in generating infographics remains unclear. Generated infographics may appear correct at first glance but contain easily overlooked issues, such as distorted data encoding or incorrect textual content. We present \system, the first benchmark for evaluating the reliability of text-to-infographic generation, comprising 600 curated test cases spanning 30 infographic types. We design an automated evaluation framework that decomposes reliability verification into atomic yes/no questions based on a taxonomy of 10 question types. We employ multimodal large language models (MLLMs) to verify each question, yielding question-level accuracy (Q-ACC) and infographic-level accuracy (I-ACC). We comprehensively evaluate 10 state-of-the-art T2I models on \system. Our systematic analysis reveals key insights for future model development: (i) a three-tier performance hierarchy with the top model achieving Q-ACC of 0.90 but I-ACC of only 0.49; (ii) data-related dimensions emerging as universal bottlenecks (e.g., Data Completeness: 0.21); and (iii) the challenge of achieving end-to-end correctness across all models. We release \system at \url{https://igen-bench.vercel.app/}.
\end{abstract}    
\section{Introduction}

Infographics are composite visual artifacts that integrate data visualizations with textual and illustrative elements, such as pictograms, thematic icons, semantic text, and metaphorical imagery~\cite{dur2014data,DBLP:journals/vldb/QinLTL20}. By integrating data with visual narratives, infographics enhance expressive power and are widely used in journalism~\cite{hamza2023importance}, education~\cite{traboco2022designing}, and business analytics~\cite{cui2019text}. Traditionally, creating high-quality infographics is labor-intensive and demands significant design expertise~\cite{cui2019text,graphimind}, often involving days of manual iteration~\cite{huang2018effect}.

Recently, advances in text-to-image (T2I) models, such as Nanobanana-Pro~\cite{nanobanana-pro} and GPT-Image~\cite{chatgpt-images}, have enabled the generation of aesthetically appealing images with complex graphics and accurate text rendering. Leveraging these advanced T2I models for automated infographic generation has become a promising direction~\cite{letchartspark,bigbenz,nanobanana-pro}. However, T2I models suffer from inherent uncertainty~\cite{uncertainty}, raising doubts about their reliability in generating structured images such as infographics. Generated charts may appear good at first glance, but often contain easily overlooked issues that can mislead users~\cite{factualitymatters,zhu2025survey}, such as distorted data encoding (e.g., incorrect bar heights) or textual errors. Such shortcomings highlight the urgent need for systematic evaluation to identify potential issues in generated infographics.

However, to the best of our knowledge, no existing benchmark is specifically designed for this task. 
First, existing infographic-related datasets focus on question answering or visual reasoning \cite{InfochartQA, mathew2021infographicvqa}, rather than on infographic generation. 
Second, there is currently no established method for evaluating the quality of infographics, and evaluation methods from related tasks do not transfer well to this setting. Prior work on text-to-image generation~\cite{scaling, photorealistic, t2i-bench} mainly evaluates prompt adherence for natural images. However, infographics require both semantic consistency between the prompt and visual elements, and accurate encoding of the underlying data values into corresponding visual representations. Moreover, existing evaluation methods for plain chart generation often rely on holistic scoring using multimodal large language models (MLLMs)~\cite{visJudge, matplotagent}. These methods offer limited interpretability and are unable to identify specific errors in the charts.

\begin{figure*}[t]
    \centering
    \includegraphics[width=\linewidth]{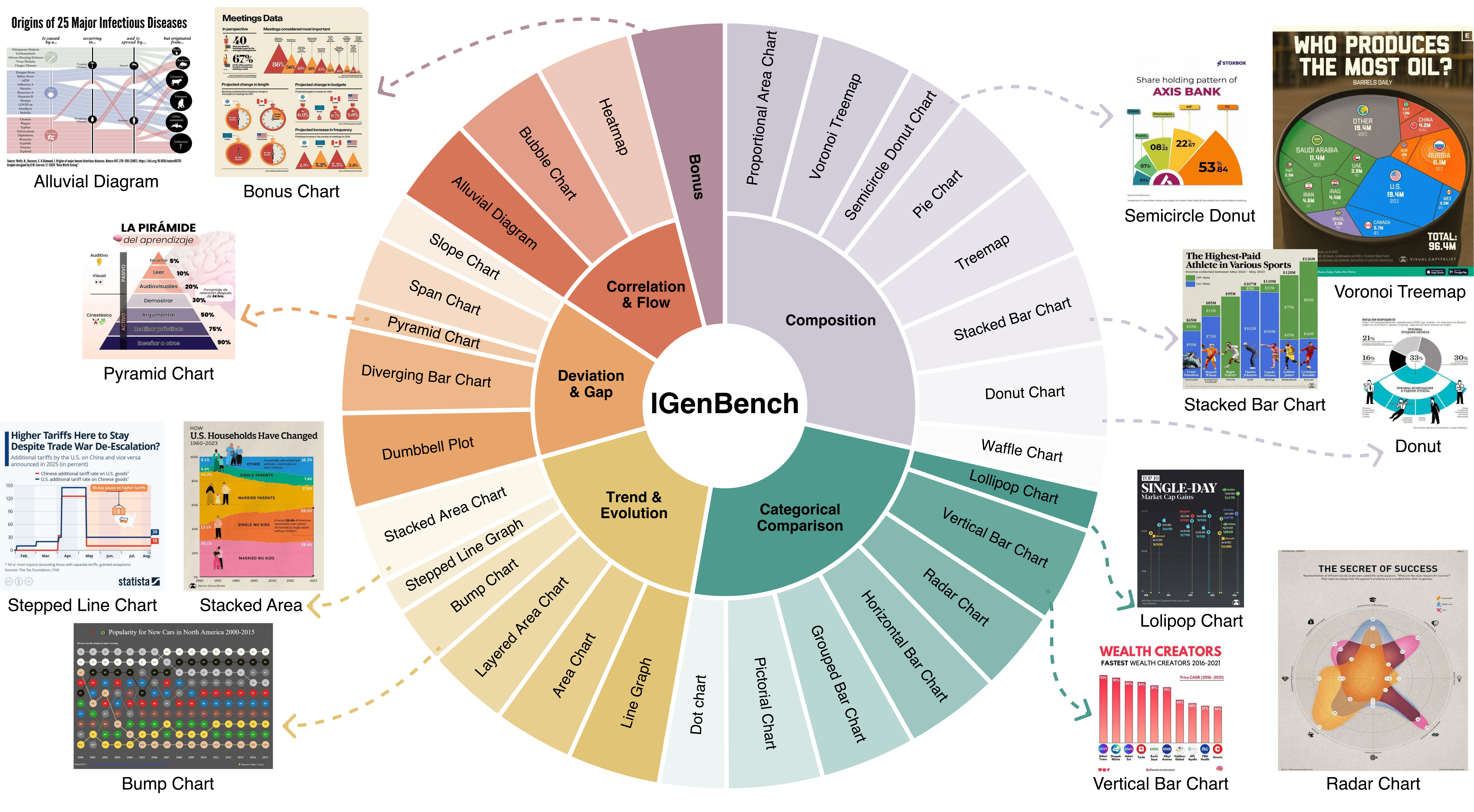}
    \caption{\system overview. The benchmark covers 30 infographic types organized into 6 high-level categories: Composition, Trend \& Evolution, Categorical Comparison, Deviation \& Gap, Correlation \& Flow, and Bonus.}
    \label{fig:pie}
    \vspace{-1em}
\end{figure*}

To address this gap, we introduce \system, the first benchmark designed to evaluate the reliability of T2I models for text-to-infographic generation. As shown in Figure \ref{fig:pie}, \system comprises 600 curated test cases spanning 30 distinct infographic types across 6 categories. 
We construct the dataset through a structured pipeline. We begin by collecting 40K real-world infographics, followed by clustering, sampling, and quality filtering to obtain 600 high-quality and diverse cases. For each case, we extract its design intent and underlying data, which are  then used to synthesize prompts that serve as self-contained specifications for infographic generation.
 
We design and implement an evaluation framework that supports automatic and interpretable assessment of generated infographics, with a focus on semantic consistency and accurate data encoding.
To enable fine-grained verification of infographic content, we first define a taxonomy of 10 question types that cover key elements, including visual components such as titles and chart types, as well as data-related aspects such as data marks. Next, we decompose the reliability verification of each generated infographic into atomic, self-contained yes/no questions, guided by the defined taxonomy. These verification questions are derived from two sources: (i) constraints explicitly specified in the prompt, and (ii) expert-informed seed dimensions, including data completeness, data ordering, and data encoding. The latter extends the coverage of question types beyond what is directly stated in the prompt. Finally, we use MLLMs to verify each question against the generated infographic, producing both question-level accuracy (Q-ACC) and infographic-level accuracy (I-ACC), where I-ACC reflects whether all specified constraints are simultaneously satisfied. This question-driven evaluation framework enables interpretable assessment of generated infographics, and shows strong agreement with human judgments (see Section~\ref{exp: alignment}).

We conduct a comprehensive evaluation of 10 state-of-the-art T2I models using \system, uncovering key limitations and trends in current infographic generation. Our experiments reveal a clear three-tier performance hierarchy. The top-tier model, Nanobanana-Pro, achieves a Q-ACC of 0.90, significantly outperforming the second-tier models, which attain Q-ACC scores between 0.55 and 0.61. The remaining models fall below a Q-ACC of 0.48. This stratification highlights fundamental capability gaps in infographic generation across current T2I models. Moreover, even the best-performing model achieves an I-ACC of only 0.49, indicating that fewer than half of its generated infographics fully satisfy all specified constraints. This underscores that current T2I models are not yet reliable for autonomous infographic generation, and that human verification and post-editing remain necessary to ensure correctness. Additionally, we identify data-related dimensions as a universal bottleneck across all models—dimensions such as Data Completeness and Data Encoding emerge as the most challenging aspects. We summarize our contributions as follows:

\begin{itemize}[noitemsep,leftmargin=10pt]
\item We present \system, the first comprehensive benchmark for evaluating infographic generation fidelity, comprising 600 curated test cases spanning 30 infographic types and covering diverse real-world generation scenarios.

\item We propose a systematic evaluation framework based on a taxonomy of 10 question types, which enables fine-grained assessment of infographic correctness using atomic yes/no verification questions. It supports both question-level accuracy (Q-ACC) and infographic-level accuracy (I-ACC) as evaluation metrics.

\item We conduct extensive experiments on 10 state-of-the-art T2I models, revealing a three-tier performance hierarchy, the challenge of achieving end-to-end correctness, and data-related aspects as a universal bottleneck.

\end{itemize}

\section{Related Work}
\label{sec:related} 

\myline{Infographics Generation.} Infographic creation traditionally requires professional designers and substantial manual effort. One direction for automation is the text-code-chart paradigm, where large language models generate visualization code (i.e., D3.js) that is then executed to produce infographics~\cite{li2025chartgalaxy, li2026deepeye}. However, these code-based approaches require extensive manual template creation and are tightly coupled to their underlying asset libraries, limiting scalability and generalizability. They also struggle to freely render visual artifacts commonly found in infographics, such as pictograms, thematic icons, and metaphorical imagery. In contrast, text-to-image (T2I) approaches have emerged as a more promising and mainstream paradigm \cite{feng2023promptmagician}. Recent works~\cite{letchartspark,visualfusion,LIDA} explore embedding semantic context into infographics for better aesthetics, while BizGen~\cite{bigbenz} advances text-rich infographics generation using layout-guided cross-attention mechanisms. With the rapid advancement of increasingly powerful T2I models like Nanobanana-Pro \cite{nanobanana-pro}, public interest in using them for infographic creation has surged. Despite this momentum, the field lacks a dedicated benchmark that systematically evaluates T2I-generated infographics.

\myline{Benchmarks.} Standardized benchmarks for infographics are lacking. Existing evaluation efforts largely focus on related but distinct tasks~\cite{wu2026autowebworld}. For general text-to-image generation, benchmarks such as PartiPrompt \cite{scaling}, DrawBench \cite{photorealistic}, TIFA \cite{tifa}, T2I-CompBench \cite{t2i-bench,t2i-bench++}, MJHQ-30K \cite{li2024playground}, and ArtiMuse \cite{caoArtiMuseFineGrainedImage2025} primarily assess prompt-following ability or visual aesthetics, often relying on vision-language models (e.g., CLIP \cite{clip} or modern multimodal LLMs \cite{qwen2.5-vl}) as automatic evaluators.  For chart and scientific visualization generation~\cite{luo2018deepeye, xie2024haichart}, benchmarks such as VisJudge-Bench~\cite{visJudge}, VIS-Shepherd~\cite{pan2025visshepherdconstructingcriticllmbased}, Vividoc \cite{tang2026demonstrating,tang2026vividoc} and MatplotBench~\cite{matplotagent} evaluate data visulization by using MLLMs to perform holistic scoring. VISEval~\cite{chen2024viseval} introduces rule-based checks to assess the legality of generated charts. StructBench~\cite{factualitymatters} and ChartMark~\cite{chen2025chartmark} further study the generation and editing of structured scientific images. For chart understanding, benchmarks such as AskChart~\cite{yang2024askchart} and ChartInsights~\cite{wu2024chartinsights} evaluate models' ability to interpret chart information. These benchmarks do not capture the unique challenges of infographic generation, which requires jointly evaluating semantic alignment and data encoding correctness. To address this gap, we propose \system, a comprehensive benchmark that enables interpretable, fine-grained assessment of  text-to-infographic generation reliability.

%However, this approach rely on well-defined chart grammars that do not generalize to infographics' diverse visual elements (decorative graphics, metaphorical imagery, flexible layouts)

\section{Dataset Construction}
\label{sec:benchmark_construction}

\begin{figure*}[t]
    \centering
    \includegraphics[width=\linewidth]{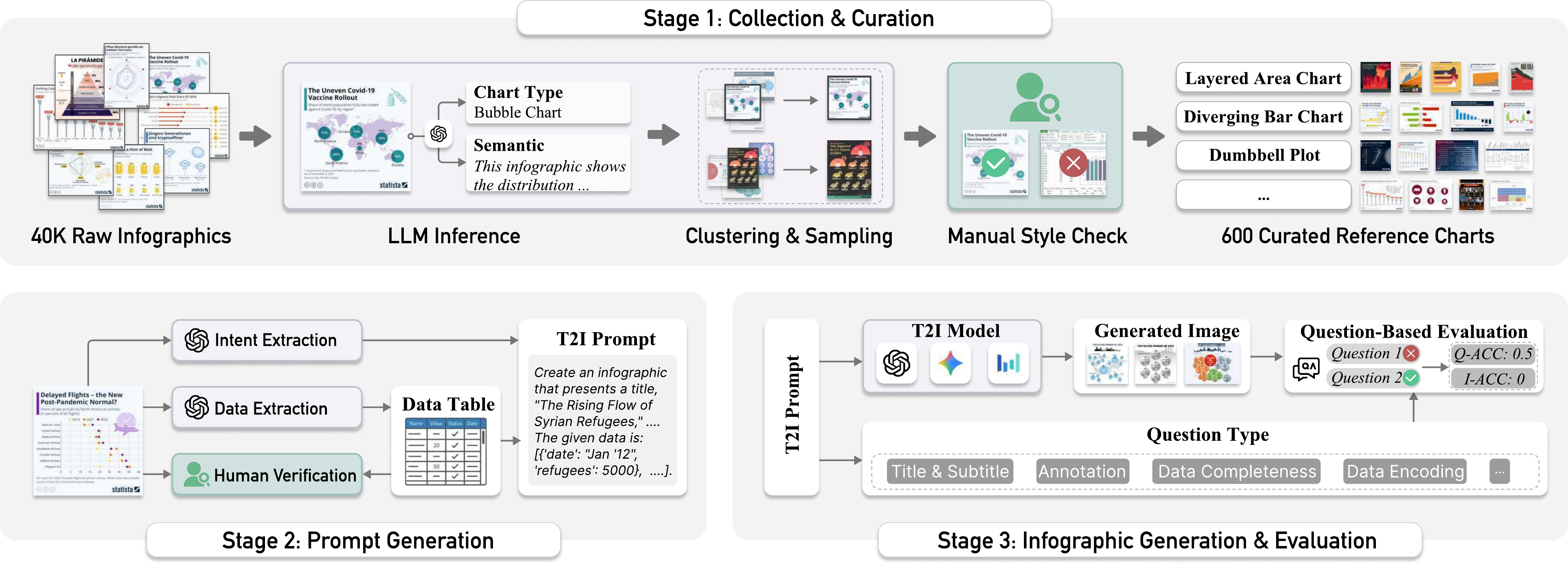}
    \caption{A three-stage pipeline for constructing \system. Stage~1 collects and curates real-world infographics. Stage~2 generates self-contained T2I prompts via human-in-the-loop intent and data extraction. Stage~3 evaluates generated infographics by decomposing prompts into atomic yes/no verification questions assessed by MLLMs.}
    \label{fig:dataset_pipeline}
    \vspace{-1em}
\end{figure*}

As shown in Stages 1 and 2 of Figure~\ref{fig:dataset_pipeline}, we construct the dataset through Collection \& Curation and Prompt Generation, which transform real-world infographic designs into infographic generation prompts. Our dataset follows two goals: (i) reflecting authentic infographic creation needs; and (ii) maximizing semantic and stylistic diversity.

\subsection{Infographic Collection \& Curation}

\myline{Real-world Infographic Collection.}
We begin by collecting a large pool of high-quality infographics from two mainstream visualization platforms, Statista\footnote{\url{https://www.statista.com/}} and Visual Capitalist\footnote{\url{https://www.visualcapitalist.com/}}, as well as the real-world portion of the ChartGalaxy dataset~\cite{li2025chartgalaxy}. For ChartGalaxy, we exclude synthetic examples to ensure that \system focuses on real-world usage scenarios.  In total, we have collected 42,315 infographic charts from these sources.

\myline{Infographics Taxonomy.}
Following prior work on chart categorization~\cite{li2025chartgalaxy} and well-established visualization taxonomies such as the Data Viz Project \cite{datavizproject}, we construct a fine-grained yet interpretable taxonomy. We refine existing taxonomies by retaining only clearly defined types and merging visually similar variants. We additionally identify multi-panel layouts as a separate \emph{bonus} category due to their unique generation challenges. This process results in a taxonomy of six high-level categories—\emph{Composition}, \emph{Trend \& Evolution}, \emph{Categorical Comparison}, \emph{Deviation \& Gap}, \emph{Correlation \& Flow}, and \emph{Bonus}—comprising 30 types in total. Detailed taxonomy construction can be found in Appendix~\ref{appendix: infographics taxonomy}.

%For instance, Voronoi treemaps and standard treemaps share the same high-level type but differ markedly in appearance and design intent. 

\myline{Clustering \& Sampling.}
Given the curated taxonomy, we first use a multimodal LLM (MLLM) to assign each infographic to its appropriate chart type and its high-level semantic description. To prevent any single semantic pattern from dominating the dataset, we perform intra-type deduplication to remove charts with substantial semantic overlap. Specifically, for each type, we apply a clustering approach to capture the distinct semantic subspaces. Within each cluster, we implement a stratified sampling procedure that filters out redundant samples based on cosine similarity to the cluster center embeddings. Finally, we manually check each  sample and remove both low-quality and visually repetitive samples to further improve the quality of the dataset. Details can be found in Appendix \ref{appendix:benchmark_construction_detail}.

\subsection{Human-in-the-Loop Prompt Generation}

After obtaining a diverse set of reference infographics, we construct generation prompts through a two-step human-in-the-loop pipeline.

\myline{Intent and Data Extraction.}
We employ an MLLM to separately extract two critical components from each infographic: (i) a structural design description that captures the layout, chart type, data encoding, text placement, and decorative elements without referencing aesthetic details such as colors, fonts, or watermarks; and (ii) the underlying data table that encodes the numerical or categorical values represented in the visualization. For the design description, we prompt the MLLM to generate a single self-contained paragraph beginning with ``Create an infographic that...'' that captures structural visual elements. For data extraction, we instruct the model to output a structured table format. Both extraction outputs undergo manual verification to ensure accuracy and completeness, correcting hallucinations or omissions introduced by the MLLM.

\myline{Prompt Synthesis.}
We fuse the verified design description and data table into a single self-contained T2I prompt. The final prompt embeds the data table directly within the design description, ending with the sentence ``The given data is: \{data\}.'' This format ensures that all necessary information—both structural intent and underlying data—is explicitly provided to the T2I model in a unified specification.

\begin{figure*}[t]
    \centering
    \includegraphics[width=\linewidth]{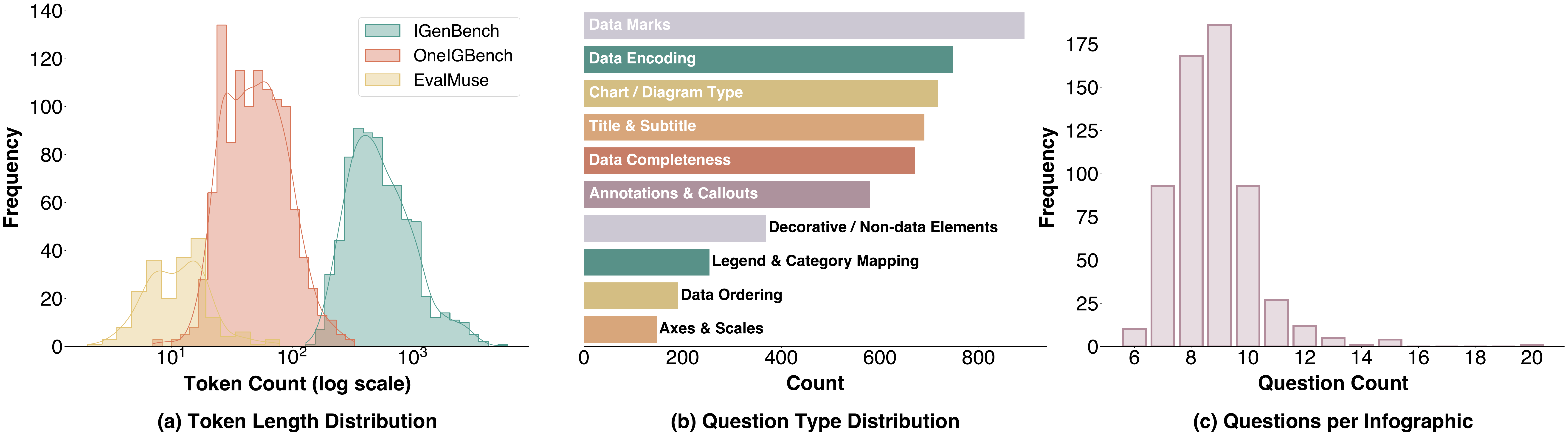}
    \caption{Statistical analysis of \system. (a) Distribution of prompt token lengths (log scale), compared with typical T2I benchmarks. (b) Distribution of the 10 question types across the benchmark. (c) Distribution of the number of verification questions per infographic.}
    \label{fig:statistical_analysis}
    \vspace{-1em}
\end{figure*}

\section{Evaluation Protocol}

As shown in Stage 3 of Figure~\ref{fig:dataset_pipeline}, we evaluate generated infographics through atomic yes/no verification questions. We then compute two complementary metrics: question-level accuracy (Q-ACC) and infographic-level accuracy (I-ACC).

\subsection{Question Set Construction}

\myline{Question Taxonomy.}
To systematically capture all critical aspects of infographic fidelity, we establish a taxonomy of 10 question types through expert consensus. Three visualization experts independently examined 300 randomly sampled infographic cases from our dataset, iteratively discussing and refining the categorization until reaching full agreement. The resulting taxonomy covers: (i) \textit{Title \& Subtitle}, verifying headings and title area text; (ii) \textit{Chart/Diagram Type}, identifying the visualization form; (iii) \textit{Decorative / Non-data Elements}, verifying icons and illustrations;  (iv) \textit{Annotations \& Callouts}, checking numeric labels and explanatory text;  (v) \textit{Axes \& Scales}, examining axis labels and tick marks; (vi) \textit{Legend \& Category Mapping}, validating color/shape/symbol keys; (vii) \textit{Data Marks}, checking primary visual objects representing data;  (viii) \textit{Data Completeness}, ensuring all required data items appear; (ix) \textit{Data Ordering}, verifying visual sequences match intended sorting logic; and (x) \textit{Data Encoding}, verifying mappings from data to visual properties. The full definitions and examples can be found in Table~\ref{tab:question_types}.

\myline{Prompt Decomposition.}
Given an input prompt $p$, we first extract a set of prompt-derived verification questions $\mathcal{Q}_p(p) = \{q_1, \ldots, q_m\}$ by decomposing the prompt into atomic constraints. Specifically, we split the prompt into individual sentences, and then use an LLM to convert each sentence into a self-contained yes/no verification question. Each $q_i$ is designed to be answerable solely by inspecting the generated infographic, without relying on external knowledge or implicit assumptions. Each question targets a specific visual or textual element based on our question taxonomy. For example, given a prompt requesting ``an infographic with title `Online dominiert den Versandhandel' and a horizontal grouped bar chart,'' we generate separate questions: one verifying the title presence (\textit{Title \& Subtitle}), and another confirming the chart type (\textit{Chart/Diagram Type}). The full prompt used for question generation is provided in Appendix~\ref{appendix:prompts}.

\paragraph{Expert-Informed Augmentation.}
Beyond explicit prompt specifications, we augment the question set with expert-informed verification questions $\mathcal{Q}_e(p) = \{q'_1, \ldots, q'_n\}$ grounded in visualization best practices. These questions instantiate three critical requirements that may be implicit in prompts: \textit{Data Completeness} (whether all required data items appear correctly), \textit{Data Ordering} (whether the visual ordering follows the specified ranking), and \textit{Data Encoding} (whether visual sizes and proportions accurately reflect the underlying data magnitudes). For each chart in the infographic, we instantiate these seed requirements into concrete, chart-specific yes/no questions following our question taxonomy. The final question set is the union: $\mathcal{Q}(p) = \mathcal{Q}_p(p) \cup \mathcal{Q}_e(p)$.

\myline{Verification.} As illustrated in Figure~\ref{fig:dataset_pipeline} (Stage 3), given a generated infographic $I$, each question $q_i \in \mathcal{Q}(p)$ is evaluated with a strict binary correctness function:
\[
\mathbb{I}(I, q_i) =
\begin{cases}
1, & \text{if } q_i \text{ is clearly satisfied in } I,\\
0, & \text{otherwise.}
\end{cases}
\]
Any ambiguity, partial satisfaction, or missing visual evidence yields a score of 0.

\subsection{Reliability Metrics}

Based on these question-level judgments, we report reliability at two aggregation levels.

\paragraph{Question-level Accuracy (Q-ACC).}
This metric measures the fraction of correctly satisfied verification questions across all evaluated infographics:
\[
\mathsf{Q\text{-}ACC} = \frac{1}{|\mathcal{Q}|}
\sum_{q_i \in \mathcal{Q}} \mathbb{I}(I, q_i),
\]
where $\mathcal{Q}$ denotes the union of all verification questions over the evaluated set.

\paragraph{Infographic-level Accuracy (I-ACC).}
This metric captures holistic correctness by measuring the fraction of infographics for which all associated verification questions are satisfied:
\[
\mathsf{I\text{-}ACC} = \frac{1}{|\mathcal{I}|}
\sum_{I \in \mathcal{I}}
\mathbb{I}\!\left(
\sum_{q_i \in \mathcal{Q}(I)} \mathbb{I}(I, q_i)
= |\mathcal{Q}(I)|
\right),
\]
where $\mathcal{Q}(I)$ denotes the question set associated with infographic $I$.

Reporting both Q-ACC and I-ACC disentangles partial correctness from complete infographic correctness: high Q-ACC indicates that a model satisfies many individual constraints, whereas high I-ACC reflects whether the model can produce an entirely correct infographic. This distinction is critical for real-world usage, where a single error may invalidate the visualization as a whole.

\subsection{Benchmark Statistics}

Figure~\ref{fig:statistical_analysis} presents key statistics of \system. As shown in Figure~\ref{fig:statistical_analysis}(a), the prompt token lengths of \system range from tens to several thousand tokens (log scale), which is 1--2 orders of magnitude longer than typical text-to-image prompts (OneIGBench \cite{changOneIGBenchOmnidimensionalNuanced2025} and EvalMuse\cite{hanEvalMuse40KReliableFineGrained2024a}). This reflects the inherent complexity of infographic generation. Figure~\ref{fig:statistical_analysis}(b) illustrates the distribution of our 10 question types across the benchmark. Figure~\ref{fig:statistical_analysis}(c) shows that most infographics are evaluated with 7--11 verification questions per case. Overall, \system contains 600 test cases, including 600 curated prompts and 5259 verification questions.

\section{Experiments}

This section reports the main experimental results of \system, including overall model performance, alignment with human evaluation, the selection of the evaluation model, and a case study. Extended experiments, including performance breakdown by chart type, data leakage analysis, and evaluator error analysis, are presented in Appendix~\ref{sec:extended_experiments}.
\label{sec:experiment}
\subsection{Experiment Setup}
We evaluated 10 leading text-to-image models on \system. Our evaluation includes 4 prominent open-source models: Qwen-Image~\cite{qwenimage}, HiDream-I1 \cite{hidream-i1}, FLUX.1-dev \cite{labs2024flux} and Z-Image-Turbo~\cite{zimage}, as well as 6 leading closed-source models: Seedream 4.5~\cite{seedream4.5}, Nanobanana~\cite{nanobanana}, Nanobanana-Pro~\cite{nanobanana-pro}, GPT-Image-1.5~\cite{chatgpt-images}, Image-01~\cite{minimax_image01}, and P-Image~\cite{pimage}. For the main evaluation, we employ Gemini-2.5-Pro \cite{gemini2_5} as the verification model to assess the generated infographics against corresponding questions.

% 1. 定义颜色
\definecolor{myblue}{RGB}{226, 176, 131}       % 用于热图的高分蓝色
\definecolor{avgpurple}{RGB}{235, 243, 242} % 专门用于 Average 行的淡紫色

% 2. 热图宏命令 \cc
\newcommand{\cc}[1]{%
    \ifdim #1pt > 0.8pt \cellcolor{myblue!40}\textbf{#1} 
    \else\ifdim #1pt > 0.6pt \cellcolor{myblue!25}#1    
    \else\ifdim #1pt > 0.4pt \cellcolor{myblue!15}#1    
    \else\ifdim #1pt > 0.2pt \cellcolor{myblue!7}#1     
    \else \color{gray!60}#1                             
    \fi\fi\fi\fi
}

\begin{table*}[t]
\centering
\small
\setlength{\tabcolsep}{2.5pt} 
\renewcommand{\arraystretch}{1.2}

\caption{Benchmark results of 10 state-of-the-art T2I models on \system. Columns show per-category Q-ACC for each of the 10 question types (sorted by average difficulty from left to right), along with overall Q-ACC and I-ACC. Darker shading indicates higher accuracy.}
\label{tab:fidelity_breakdown_models}

\resizebox{\textwidth}{!}{%
\begin{tabular}{l cccccccccc : cc}
\toprule
\multirow{2}{*}{\textbf{Model}} &
\multicolumn{10}{c}{\textbf{Question Type} (sorted by average)} &
\multicolumn{2}{c}{\textbf{Overall}} \\
\cmidrule(lr){2-11} \cmidrule(lr){12-13}
 & \scriptsize\faCalculator~Comp.
 & \scriptsize\faCode~Enc.
 & \scriptsize\faChartLine~Order
 & \scriptsize\faMarker~Marks
 & \scriptsize\faEdit~Anno.
 & \scriptsize\faRuler~Axes
 & \scriptsize\faListUl~Leg.
 & \scriptsize\faChartPie~Chart
 & \scriptsize\faHeading~Title
 & \scriptsize\faGem~Deco.
 & Q-ACC$\uparrow$ & I-ACC$\uparrow$ \\
\midrule

Nanobanana-Pro
& \cc{0.84} & \cc{0.86} & \cc{0.90} & \cc{0.87} & \cc{0.93} & \cc{0.93} & \cc{0.96} & \cc{0.92} & \cc{0.98} & \cc{0.94}
& \cc{0.90} & \cc{0.49} \\

Seedream-4.5
& \cc{0.34} & \cc{0.37} & \cc{0.47} & \cc{0.48} & \cc{0.70} & \cc{0.70} & \cc{0.81} & \cc{0.68} & \cc{0.95} & \cc{0.84}
& \cc{0.61} & \cc{0.06} \\

GPT-Image-1.5
& \cc{0.38} & \cc{0.48} & \cc{0.44} & \cc{0.57} & \cc{0.50} & \cc{0.54} & \cc{0.57} & \cc{0.68} & \cc{0.60} & \cc{0.80}
& \cc{0.55} & \cc{0.12} \\

\hdashline[1pt/2pt] \noalign{\smallskip}

Nanobanana
& \cc{0.18} & \cc{0.31} & \cc{0.27} & \cc{0.44} & \cc{0.54} & \cc{0.57} & \cc{0.52} & \cc{0.60} & \cc{0.65} & \cc{0.81}
& \cc{0.48} & \cc{0.02} \\

Qwen-Image
& \cc{0.10} & \cc{0.13} & \cc{0.19} & \cc{0.29} & \cc{0.43} & \cc{0.37} & \cc{0.51} & \cc{0.48} & \cc{0.56} & \cc{0.78}
& \cc{0.36} & \cc{0.01} \\

Z-Image-Turbo
& \cc{0.10} & \cc{0.16} & \cc{0.16} & \cc{0.25} & \cc{0.38} & \cc{0.31} & \cc{0.58} & \cc{0.42} & \cc{0.61} & \cc{0.73}
& \cc{0.35} & \cc{0.00} \\

P-Image
& \cc{0.08} & \cc{0.15} & \cc{0.19} & \cc{0.27} & \cc{0.36} & \cc{0.28} & \cc{0.54} & \cc{0.43} & \cc{0.58} & \cc{0.68}
& \cc{0.34} & \cc{0.00} \\

Image-01
& \cc{0.01} & \cc{0.05} & \cc{0.04} & \cc{0.10} & \cc{0.10} & \cc{0.14} & \cc{0.03} & \cc{0.22} & \cc{0.14} & \cc{0.47}
& \cc{0.13} & \cc{0.00} \\

HIDream-I1
& \cc{0.01} & \cc{0.03} & \cc{0.03} & \cc{0.10} & \cc{0.07} & \cc{0.14} & \cc{0.10} & \cc{0.26} & \cc{0.19} & \cc{0.20}
& \cc{0.11} & \cc{0.00} \\

FLUX.1-dev
& \cc{0.00} & \cc{0.03} & \cc{0.01} & \cc{0.08} & \cc{0.06} & \cc{0.06} & \cc{0.01} & \cc{0.24} & \cc{0.09} & \cc{0.39}
& \cc{0.10} & \cc{0.00} \\

\midrule
\rowcolor{avgpurple}
\textbf{Average}
& \textbf{0.21} & \textbf{0.26} & \textbf{0.27} & \textbf{0.35} & \textbf{0.40} & \textbf{0.40}
& \textbf{0.46} & \textbf{0.49} & \textbf{0.54} & \textbf{0.66}
& \textbf{0.39} & \textbf{0.07} \\
\bottomrule
\end{tabular}
}
\vspace{-1em}
\end{table*}

\subsection{Main Results}

\myline{Tiered performance reveals fundamental capability gaps.}
As shown in Table~\ref{tab:fidelity_breakdown_models}, the experimental results reveal a clear three-tier performance hierarchy among the evaluated T2I models for infographic generation. The top-tier model, Nanobanana-Pro, achieves a Q-ACC of 0.90, significantly outperforming all other models. The second tier, consisting of Seedream-4.5 and GPT-Image-1.5, demonstrates moderate performance with Q-ACC scores of 0.61 and 0.55, respectively. The remaining models fall into a third tier with Q-ACC below 0.5, indicating fundamental difficulties in generating accurate infographics. This stratification is consistent across nearly all evaluation dimensions. For instance, in the Annotations \& Callouts dimension, the performance gap between the top model (0.93) and the average (0.40) exceeds 0.50 points. Moreover, the average Q-ACC across all models is only 0.39, highlighting the difficulty of reliable infographic generation.

\myline{Data fidelity remains the primary bottleneck.}
As shown in Table~\ref{tab:fidelity_breakdown_models}, data-related evaluation dimensions consistently emerge as the most challenging aspects of infographic generation across all models. Data Completeness shows the lowest average performance at 0.21, followed by Ordering (0.27) and Data Encoding (0.26). Even the best-performing model, Nanobanana-Pro, achieves only 0.84 for Data Completeness and 0.86 for Data Encoding, leaving room for improvement.

This pattern reveals a fundamental limitation of current T2I models: while they excel at generating aesthetically pleasing visual layouts, chart types, titles, and decorative elements (average scores of 0.49, 0.54, and 0.46, respectively), they struggle with the precise rendering and faithful encoding of underlying data values—a capability that remains underdeveloped in current T2I models optimized primarily for natural image generation.

\myline{High Q-ACC does not imply reliable infographics.}
As shown in Table~\ref{tab:fidelity_breakdown_models}, there exists a dramatic gap between Q-ACC and I-ACC across all models, with I-ACC consistently much lower. The best-performing model achieves a Q-ACC of 0.90 but only 0.49 I-ACC. This difference is larger for second- and third-tier models: Seedream-4.5 and GPT-Image-1.5 drop from 0.61 and 0.55 Q-ACC to only 0.06 and 0.12 I-ACC, respectively, while most other models have I-ACC scores near or at zero.

The low I-ACC scores indicate that current T2I models show a ``long-tail'' failure mode: although they may correctly generate many aspects of an infographic, they often fail on at least one or two critical dimensions, which prevents end-to-end correctness. This suggests that for real-world use, particularly in high-stakes domains such as business analytics or education, where information accuracy is essential, current T2I models cannot yet be trusted to autonomously generate reliable infographics, even when their component-level metrics appear promising.

\begin{figure}[htbp]
    \centering
    \includegraphics[width=\columnwidth]{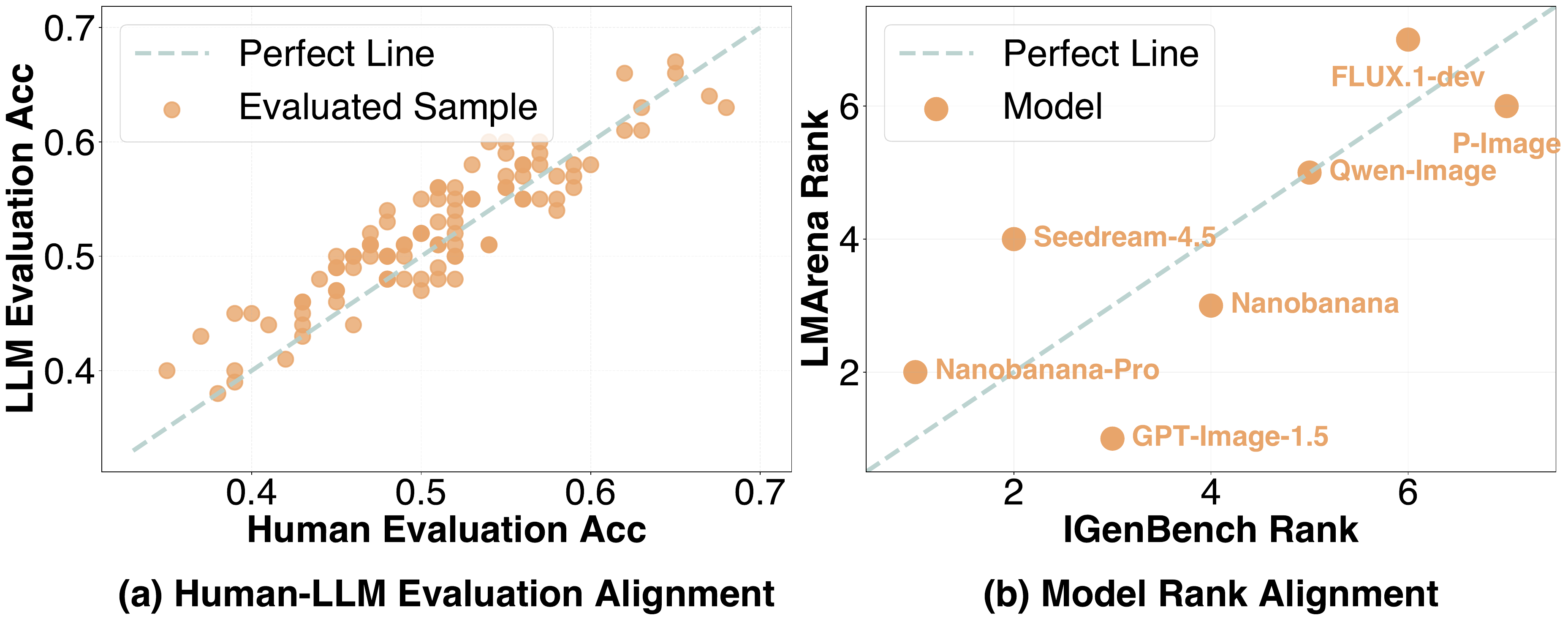}
    \caption{(a) Correlation between automatic evaluation (Gemini-2.5-Pro) and human judgments across 100 bootstrap samples of 25 questions each, showing strong Pearson correlation ($r=0.90$). (b) Comparison of model rankings between \system and LMArena, with Spearman $\rho=0.78$.}
    \label{fig:alignment}
    \vspace{-1em}
\end{figure}

\begin{figure*}[t]
  \centering
  \includegraphics[width=\linewidth]{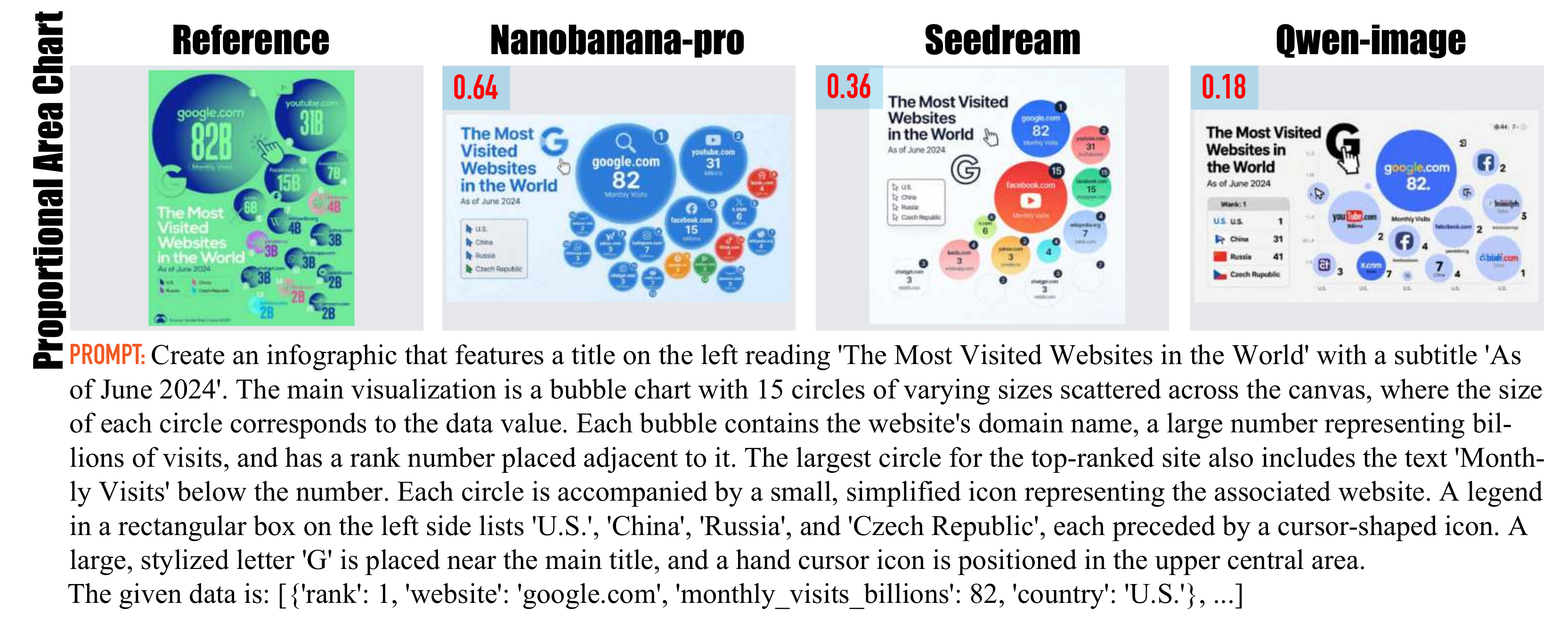}
  \caption{Case of Proportional Area Chart. The prompt specifies 15 bubbles with sizes proportional to website visit counts. Nanobanana-Pro generates 16 bubbles with some color encoding errors; Seedream and Qwen-Image exhibit more severe issues including incorrect ranking order and garbled text.}
  \label{fig:case_study_proportional_area_chart}
  \vspace{-1em}
\end{figure*}

\subsection{Alignment with Human Evaluation and LMArena}
\label{exp: alignment}
\myline{Automatic evaluation aligns strongly with human judgment.} Following prior work~\cite{matplotagent}, we assess the reliability of automatic evaluation by measuring its correlation with human judgments. Specifically, we use Gemini-2.5-Pro as the evaluator and recruit expert annotators to assess the same set of generated infographics. For each T2I model, we iteratively sample subsets of 25 examples from its generated outputs and compute average scores from both automatic and human evaluation. This process is repeated 100 times, yielding 100 data points (Figure \ref{fig:alignment}a) for each evaluation type: $A = \{a_1, \cdots, a_{100}\}$ and $H = \{h_1, \cdots, h_{100}\}$, where $a_i$ and $h_i$ denote the average automatic and human scores on the $i$-th sampled subset, respectively. We obtain a Pearson correlation of $r=0.90$ with $p=7.54\times10^{-37}$. Given that $r > 0.8$ and $p < 0.05$, we conclude that automatic evaluation scores strongly correlate with human judgments, validating the reliability of \system.

\myline{IGenBench rankings correlate with but diverge from natural image benchmarks.} To further validate the characteristics of our benchmark, we compare model rankings on \system with those on LMArena~\cite{lmarena_ai}, a widely-used arena for natural image generation. We select 7 models that are evaluated in both \system and LMArena. We calculate the Spearman correlation between the two rankings and obtain $\rho=0.78$ with $p=0.04$. This moderate-to-strong positive correlation suggests that models with strong performance on natural image generation often also perform well on infographic generation.

At the same time, the observed ranking differences highlight challenges that are specific to infographic generation. For example, Seedream-4.5 ranks fourth on LMArena but places second on \system, whereas GPT-Image-1.5 ranks first on LMArena but falls to third on \system. These differences suggest that infographic generation requires more than photorealistic rendering, including accurate data encoding, compliance with structured layouts, and correct semantic alignment between visual elements and underlying data.

\subsection{Selection of Evaluation Model}
To identify an appropriate evaluator for the infographic assessment task, we examine the alignment between different MLLMs and human judgments. Following the setup in Experiment~\ref{exp: alignment}, we evaluate 9 leading open-source MLLMs (Llama-4-Maverick \cite{meta_llama4_multimodal}, Mistral-Small-3.2-24b \cite{mistral_small_3_2}, Qwen3-VL-8b \cite{yang2025qwen3technicalreport}, Qwen3-VL-32b, Qwen2.5-VL-72b \cite{qwen2.5-vl}, GLM-4.5v \cite{glm_4_5_multimodal}, Gemma-3-27b \cite{google_gemma3}, Pixtral-12b \cite{agrawal2024pixtral12b}, and InternVL3-78b \cite{zhu2025internvl3exploringadvancedtraining}) and 3 closed-source models (Gemini-2.5-Pro \cite{gemini2_5}, GPT-5-mini \cite{openai_gpt5}, and Grok-4.1 \cite{xai_grok4_1}) by computing Pearson correlation with human evaluations on the same set of verification questions.

\begin{figure}[h]
    \centering
    \includegraphics[width=\columnwidth]{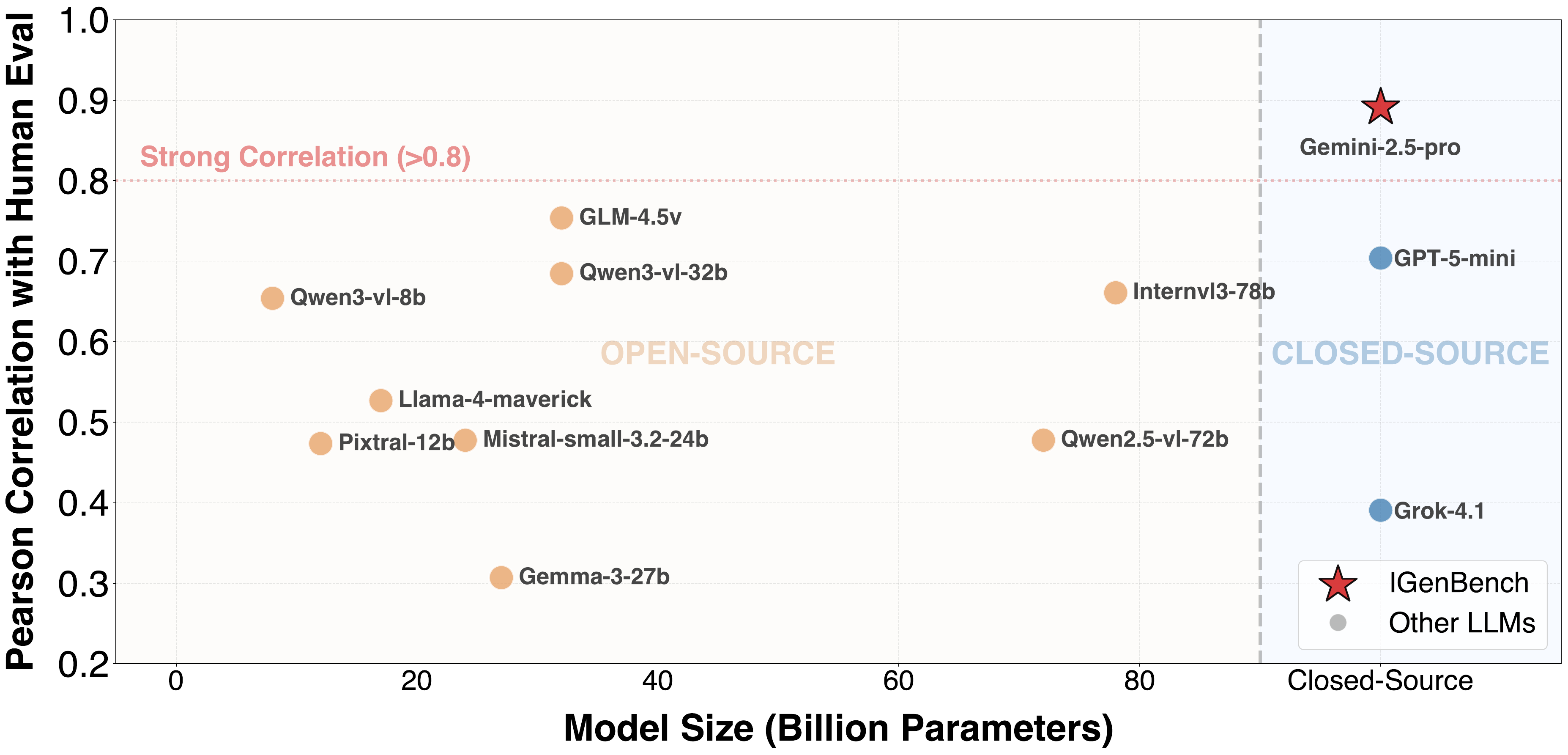}
    \caption{Pearson correlation between different MLLMs' automatic scores and human judgments. Gemini-2.5-Pro achieves the highest correlation ($r=0.90$), the only model surpassing the strong correlation threshold of 0.8.}
    \label{fig:corr}
    \vspace{-1em}
\end{figure}

As shown in Figure~\ref{fig:corr}, Gemini-2.5-Pro achieves a Pearson correlation of 0.90, making it the only model surpassing the strong correlation threshold of 0.8. GPT-5-mini (0.70) and GLM-4.5v (0.75) show moderate alignment, while most open-source models exhibit substantially lower correlations, with some falling below 0.5. Based on these results, we select Gemini-2.5-Pro as our automated evaluator throughout all experiments.

\subsection{Case Study}
As shown in Figure~\ref{fig:case_study_proportional_area_chart}, we present a representative case involving a proportional area chart displaying ``The Most Visited Websites in the World.'' The prompt specifies 15 bubbles with sizes proportional to visit counts. Nanobanana-Pro generates 16 bubbles instead of the required 15 and exhibits incorrect color encoding for certain data points. Seedream and Qwen-Image produce more severe errors, such as incorrect ranking order and garbled text in annotations. This case illustrates the difficulty of simultaneously satisfying multiple fine-grained constraints in infographic generation. Additional representative cases are provided in Appendix~\ref{appendix:full_case}, and the complete case gallery is available on the project website at \url{https://igen-bench.vercel.app/}.

\section{Conclusion}
We present \system, the first benchmark for evaluating the reliability of text-to-infographic generation. Through 600 curated test cases spanning 30 infographic types and a question-driven evaluation framework, we assess current T2I models' capability to generate reliable infographics. Our evaluation of 10 state-of-the-art models reveals critical limitations of current T2I models and key insights for future model development.
\section{Limitation}

This work focuses on evaluating the reliability of text-to-infographic generation, with an emphasis on semantic consistency between the prompt and visual elements, as well as accurate encoding of the underlying data values into corresponding visual representations. As a result, \system does not assess broader evaluation dimensions such as communicative effectiveness (whether the infographic successfully conveys its intended message), accessibility (e.g., colorblind-friendliness), or visual aesthetics (e.g., layout organization and stylistic creativity). We view reliability as a foundational prerequisite for these dimensions: accessibility presupposes correct visual encoding, and communicative effectiveness presupposes faithful data representation. Moreover, these dimensions require fundamentally different evaluation methodologies, such as user studies for communicative effectiveness and perceptual modeling for accessibility, which fall outside the scope of this work. We leave the systematic assessment of these complementary dimensions to future work. In addition, due to the high monetary cost of large-scale evaluation, we only include a selected set of representative state-of-the-art models. We view \system as a living benchmark and plan to continuously incorporate more models.
\section{Ethical Considerations}

All infographic images used in this work are publicly available and obtained from open-access sources. To address potential copyright issues, we will not redistribute the original images and will only release their URLs as part of the dataset. We manually reviewed all collected images to verify that they do not contain harmful, illegal, or sensitive content. Some infographics include references to humans. These cases are limited to public figures appearing in widely distributed media content, such as news articles or public reports. The dataset does not involve private individuals or personal user data, and it does not include information intended for restricted or private use. As this work focuses on benchmarking and evaluation rather than deploying or enabling new generative systems, we do not foresee significant ethical or societal risks. 

\section*{Acknowledgments}
This work was supported in part by the National Natural Science Foundation of China (62132017, 62502430, 62402409) and the Zhejiang Provincial Natural Science Foundation of China (LQ26F020004).

\bibliography{custom}
\newpage
\appendix
\clearpage
\section*{Appendix Overview}

This appendix provides supplementary materials supporting the \system benchmark. Section~\ref{section:llm_usage} describes the use of large language models (LLMs) in this work. Section \ref{section:human_participation} reports human involvement in this study. Section~\ref{appendix:benchmark_construction_detail} details the dataset construction process, including clustering and sampling, manual quality filtering, as well as the resultant infographic and question taxonomies. Section~\ref{sec:extended_experiments} reports additional analyses beyond the main text, including a study of potential data leakage, an error analysis of the automated evaluation, an evaluator bias investigation across multiple providers, a reference-generation visual similarity analysis, a prompt length sensitivity study, and a question type independence analysis. Section~\ref{appendix:full_case} presents representative case studies, with the complete case gallery available on the project website. Section~\ref{appendix:prompts} documents all prompts used throughout the pipeline.

\section{LLM Usage}
\label{section:llm_usage}
We used Claude-4.5-Sonnet for English grammar polishing of the paper. During dataset construction, we used MLLMs to assist with synthesizing text-to-infographic generation prompts, as part of a human-in-the-loop process rather than as autonomous decision makers. During evaluation, we used MLLMs to automatically answer atomic yes/no verification questions against generated infographics.

\section{Human Participation}
\label{section:human_participation}
Human involvement in this work included expert discussions during benchmark design and targeted annotation during dataset curation and evaluation. The taxonomy of question types was developed through iterative discussions among three coauthors with domain expertise to ensure coverage of key infographic elements. During dataset construction, human annotators filtered low-quality infographics to maintain dataset quality. For human evaluation, three undergraduate students majoring in computer science were recruited locally and compensated according to local wage standards. They were asked to answer the same atomic yes/no verification questions used in the automatic evaluation. The process was guided by an instruction, which is shown in Figure~\ref{fig:instruction_human}. The study did not involve sensitive personal data, and participants were not exposed to harmful or risky content.

\section{Benchmark Construction Detail}
\label{appendix:benchmark_construction_detail}

\subsection{Clustering \& Sampling Algorithm.} 
As shown in Algorithm~\ref{alg:per-type-cluster-then-sample}, we employ a stratified cluster-then-sample strategy to ensure diversity and representativeness across different chart types. For each chart type $c$, we first apply $k$-means clustering on the semantic embeddings ${e_i = f(x_i)}$ extracted from the infographic descriptions or visual content, partitioning samples into $C$ clusters. Within each cluster, we select the medoid (the sample closest to the cluster centroid) to capture the most representative instance, and optionally sample additional diverse instances. This process is repeated until $K$ samples are selected for each chart type. This approach yields a balanced and comprehensive benchmark that reflects the full spectrum of real-world infographic complexity. $K$ and $C$ are set to 5 and 10, respectively.

\begin{algorithm}[h]
  \caption{Per-Type Sampling}
  \label{alg:per-type-cluster-then-sample}
  \KwIn{samples $D = \{(x_i, t_i)\}$, embeddings $e_i = f(x_i)$, samples per type $K$, clusters $C$}
  \KwOut{selected sample indices $S$}
  $S \gets \varnothing$\;
  
  \ForEach{chart type $c$}{
      Run $k$-means with $C$ clusters on $\{ e_i \mid t_i = c \}$\;
      Let clusters be $\{ I_1, \dots, I_C \}$ with centroids $\{\mu_1, \dots, \mu_C\}$\;
      $S_c \gets \varnothing$\;
      
      \For{$j=1$ \KwTo $C$}{
          \If{$|S_c| \ge K$}{\textbf{break}\;}
          $i^\star \gets \arg\min_{i \in I_j} \mathrm{dist}(e_i, \mu_j)$ \tcp*{select medoid}
          $S_c \gets S_c \cup \{ i^\star \}$\;
      }
      $S \gets S \cup S_c$\;
  }
  \Return $S$\;
\end{algorithm}

\subsection{Manual Style Check of Dataset} 
\begin{figure*}[htbp]
\centering
\includegraphics[width=1.0\textwidth]{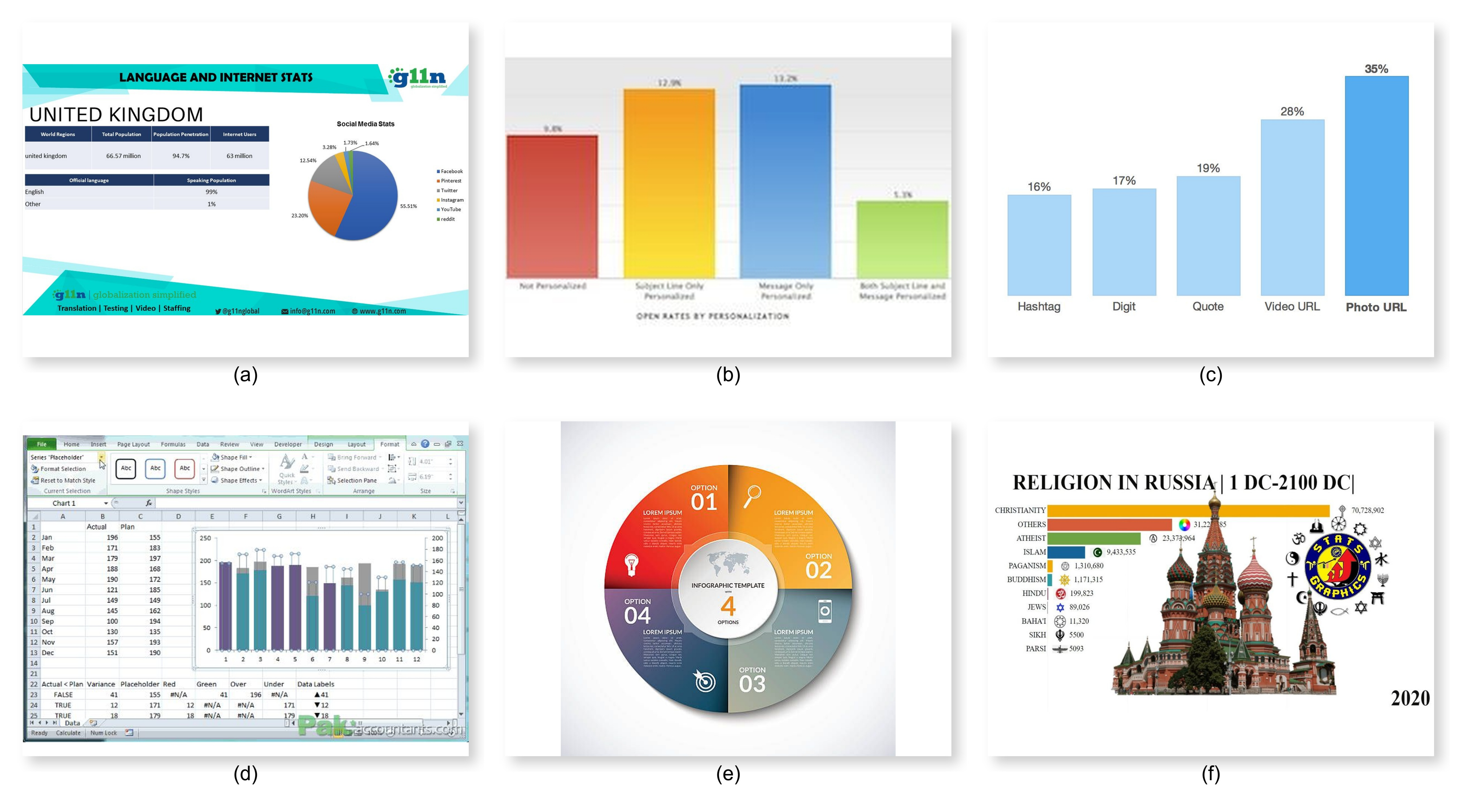}
\caption[Bad cases in raw dataset]{Representative bad cases identified during manual quality review. Examples of samples excluded from \system: (a) semantically unrelated charts combined without coherent theme; (b) low-resolution and pixelated images compromising legibility; (c) pure chart elements lacking infographic narrative features; (d) application interface screenshots rather than standalone visualizations; (e) empty templates without actual data content; (f) visual clutter with non-essential elements obscuring data presentation.}
\label{fig:bad_cases}
\end{figure*}

Following the clustering and sampling process, we conducted a rigorous manual style check to ensure the quality and suitability of selected samples for benchmark evaluation. During this review, we identified and filtered out samples that did not meet our quality standards. These problematic cases generally fall into categories such as irrelevance, poor legibility, or lack of informational content. Figure~\ref{fig:bad_cases} illustrates six representative types of bad cases identified during manual inspection.

As seen in Figure~\ref{fig:bad_cases}(a), some images combine charts with no semantic relevance to one another, making them unsuitable as coherent infographics.
Figure~\ref{fig:bad_cases}(b) shows examples where text and visual elements are pixelated and low-resolution, severely compromising legibility.
Some samples, such as Figure~\ref{fig:bad_cases}(c), contain only pure chart elements without the narrative or design features that characterize infographics.
Figure~\ref{fig:bad_cases}(d) depicts a screenshot of a computer application interface rather than a standalone visualization.
Figure~\ref{fig:bad_cases}(e) represents template images containing layout structures but lacking actual data information.
In (f), visual elements interfere with data presentation by partially obscuring labels and complicating chart readability. All samples exhibiting these issues were excluded from the final benchmark to ensure high-quality evaluation.

\subsection{Infographics Taxonomy.} 
\label{appendix: infographics taxonomy}

We aim to ensure the benchmark is structured around the major chart types commonly recognized in visualization taxonomies. Many existing efforts categorize charts into roughly ten coarse-grained families \cite{xu2023chartbench,han2023chartllamamultimodalllmchart, xia2025chartxchartvlmversatile}; however, such coarse categories may mask significant stylistic variation. ChartGalaxy~\cite{li2025chartgalaxy} provides over seventy fine-grained real-world categories, yet several are visually similar to one another or not clearly defined in widely used visualization taxonomies such as the Data Viz Project \cite{datavizproject}. To construct a fine-grained and interpretable taxonomy, we retain only the ChartGalaxy types that appear in existing public taxonomies and merge the remaining types into the most visually similar categories. We additionally identify infographic charts that contain multi-panel (multi-chart) layouts, which are particularly difficult for both classification and subsequent generation; these are grouped separately into a \emph{bonus} category. This process results in a fine-grained taxonomy consisting of six high-level categories with 30 types in total:

\textbf{Category 1 (Composition):} Pie Chart, Donut Chart, Semicircle Donut, Stacked Bar, Treemap, Voronoi Treemap, Waffle Chart, Proportional Area.

\textbf{Category 2 (Categorical Comparison):} Vertical Bar, Horizontal Bar, Grouped Bar, Lollipop Chart, Radar Chart, Pictorial Chart, Dot Chart.

\textbf{Category 3 (Trend \& Evolution):} Line Graph, Stepped Line, Area Chart, Layered Area, Stacked Area, Bump Chart.

\textbf{Category 4 (Deviation \& Gap):} Diverging Bar, Pyramid Chart, Dumbbell Plot, Slope Chart, Span Chart.

\textbf{Category 5 (Correlation \& Flow):} Bubble Chart, Heatmap, Alluvial Diagram.

\textbf{Category 6 (Bonus):} Multi-panel and multi-chart layouts that combine multiple visualization types within a single infographic.

\begin{table*}[htbp]
    \centering
    \caption{Question Types, Descriptions, and Examples}
    \label{tab:question_types}
    \renewcommand{\arraystretch}{1.3} % 减少行高避免溢出
    
    % 定义列格式：
    % l: 左对齐 (第一列)
    % X: 自动换行并填充剩余空间 (第二、三列)
    \begin{tabularx}{\textwidth}{@{}p{3cm} >{\RaggedRight\arraybackslash}X >{\RaggedRight\arraybackslash}X@{}}
        \toprule
        \textbf{Question Type} & \textbf{Definition} & \textbf{Example} \\
        \midrule
        
        Title \& Subtitle & 
        Questions focusing on the main heading, sub-headings, or the text content of the title area. & 
        ``Does the infographic feature the title `Years in MLS and Average Game Attendance, 2017' at the top left?'' \\
        \cmidrule{1-3}
        
        Chart / Diagram Type & 
        Questions identifying the specific classification, style, or overall form of the visualization. & 
        ``Is the main visual a single filled area chart showing a rising trend over time, plotted against a grid of horizontal dotted lines?' \\
        \cmidrule{1-3}

        Decorative / Non-data Elements & 
        Questions about icons, illustrations, or artistic elements that do not directly encode data. & 
        ``Is there a cartoon robot holding money sitting on the lower data line on the right?'' \\
        \cmidrule{1-3}

        Annotations \& Callouts & 
        Questions about specific numeric labels, explanatory text, or callouts not part of axes. & 
        ``Is there a separate box in the bottom right corner that presents the text `U.S. Overall' along with the national average growth rate?'' \\
        \cmidrule{1-3}

        Axes \& Scales & 
        Questions about axes, tick marks, gridlines, ranges, or scale labels. & 
        ``Is the vertical y-axis on the left labeled with the percentage values `+120\%', `+80\%', `+40\%', `0\%', and `-40\%'?'' \\
        \cmidrule{1-3}
        
        Legend \& Category Mapping & 
        Questions involving the key that explains how colors, shapes, or symbols correspond to categories. & 
        ``Is a legend indicating two series, `2011' and `2012', located above the chart area?'' \\
        \cmidrule{1-3}
        
        Data Marks & 
        Questions regarding the primary visual objects (e.g., bars, points, regions) that directly represent data. & 
        ``Is each of the 10 highlighted states on the map marked with a numbered circle indicating its rank from 1 to 10?' \\
        \cmidrule{1-3}

        Data Completeness & 
        Questions verifying whether the visualization includes all expected data points, categories, or specific entities without omission or extraneous additions. & 
        ``Does the treemap chart display rectangles for exactly nine brands: Tesla, BYD, AION, SGMW, Volkswagen, BMW, Hyundai, MG, and KIA?'' \\
        \cmidrule{1-3}
        
        Data Ordering & 
        Questions verifying that the visual sequence of elements matches the intended sorting logic (e.g., descending, ascending) described in the design. & 
        ``Is the entire chart sorted in descending order by average attendance?'' \\   
        \cmidrule{1-3}
        
        Data Encoding & 
        Questions about how data values are mapped to visual properties like size, color, or position. & 
        ``Are the relative areas of the polygonal cells proportional to their numerical values, such that the cells for China (814) and USA (800) are the largest and nearly equal in size, followed by the `Other' cell (327) which appears slightly larger than the India cell (271), and the cells for Sweden (25), Spain (27), and Israel (29) are among the smallest?'' \\
        
        %\bottomrule
    \end{tabularx}
\end{table*}

\subsection{Question Taxonomy} \system evaluates infographic generation quality through a comprehensive set of verification questions that systematically assess whether generated outputs faithfully reproduce all specified design elements and data content from the input prompts. As described in the main text, we established a taxonomy of 10 question types through expert consensus, where three visualization experts independently examined 300 randomly sampled infographic cases, iteratively refining the categorization until reaching full agreement.

Table~\ref{tab:question_types} provides detailed definitions and representative examples for each question type in our taxonomy. These categories comprehensively cover both data-driven elements (e.g., \textit{Data Marks}, \textit{Data Encoding}) and design-oriented components (e.g., \textit{Title \& Subtitle}, \textit{Legend \& Category Mapping}, \textit{Decorative/Non-data Elements}). Additionally, meta-level properties such as \textit{Data Completeness} and \textit{Ordering} ensure that generated infographics not only include correct elements but also maintain proper structural relationships. By decomposing infographic fidelity into these fine-grained categories, \system enables precise diagnosis of model strengths and weaknesses across different aspects of infographic generation.

\section{Extended Experiments}
\label{sec:extended_experiments}

\subsection{Performance Breakdown on Chart Type}

\begin{figure*}[!tp]
  \centering
  \includegraphics[width=\linewidth]{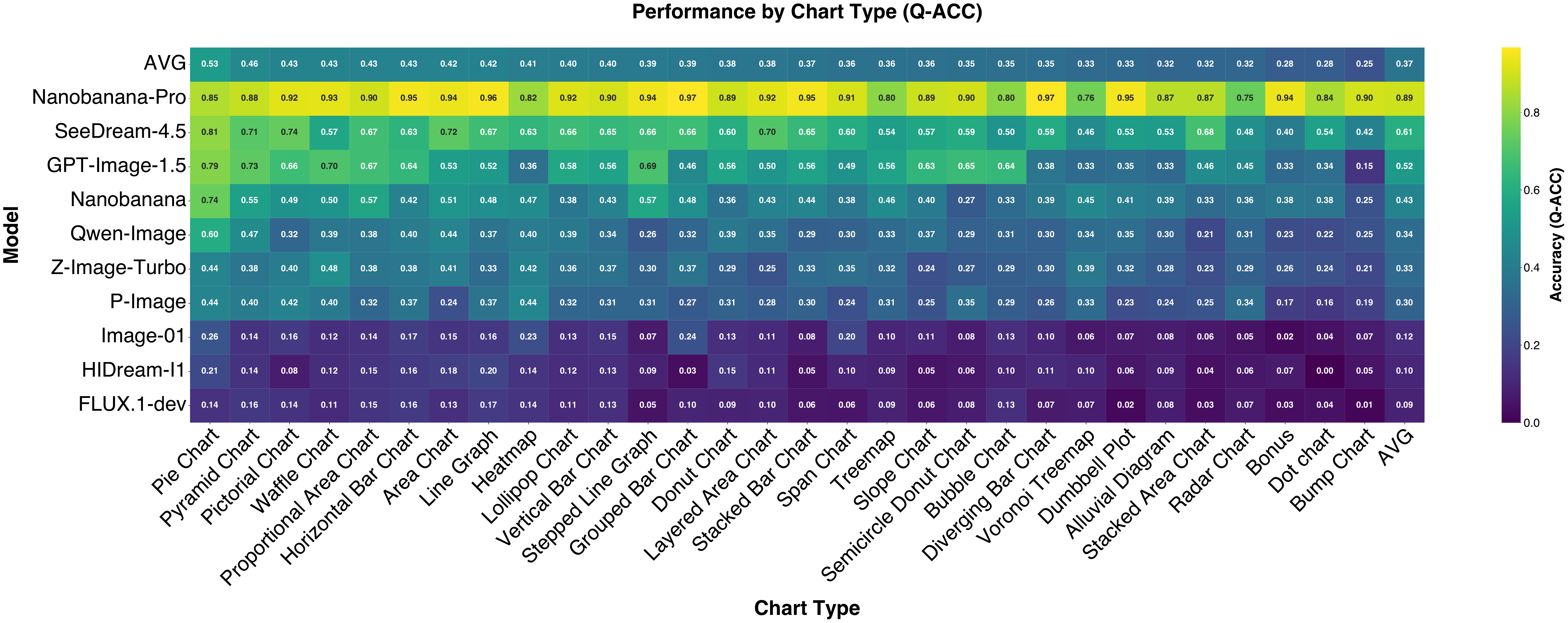}
  \caption{Performance breakdown by chart type. Q-ACC of all 10 models across 30 chart types. Common visualizations (e.g., Pie Chart, Bar Chart) achieve higher accuracy, while structurally complex types (e.g., Alluvial Diagram, Bump Chart) show much lower performance.}
  \label{fig:acc_by_chart_type}
\end{figure*}

Figure~\ref{fig:acc_by_chart_type} reports the Q-ACC of all models across 30 chart types. Two key observations emerge.

\myline{Model performance varies substantially across chart types, reflecting chart-specific difficulty.}
Across all models, Q-ACC differs markedly by chart type. Canonical visualizations with simple and widely used grammars, such as Pie Chart, Bar Chart, and Line Graph, achieve higher accuracy. In contrast, charts with complex layouts or unconventional encodings, including Alluvial Diagram, Radar Chart, Voronoi Treemap, and Bump Chart, show much lower performance. For example, on Pie Chart, the average Q-ACC across models is 0.53, while on Bump Chart it is only 0.25. This variation indicates that infographic generation has different levels of difficulty across chart types, mainly due to differences in structural complexity, spatial constraints, and the precision required to correctly align data elements.

\myline{Model rankings are largely stable across chart types.}
To quantify ranking consistency, we compute the Spearman rank correlation between model rankings across all pairs of chart types based on Q-ACC. We observe that the average Spearman rank correlation is 0.92, indicating a high degree of consistency in relative model ordering. Nonetheless, occasional deviations are observed, with the minimum correlation dropping to approximately 0.68. These fluctuations predominantly arise in comparisons involving structurally complex or less common chart types. For example, on Pie Chart, SeedDream-4.5 and GPT-Image-1.5 rank fourth (0.79),  whereas on Bump Chart, the same model drops to seventh (0.15). 

\subsection{Potential Data Leakage} 
\begin{figure}[!htbp]
    \centering
    \includegraphics[width=\columnwidth]{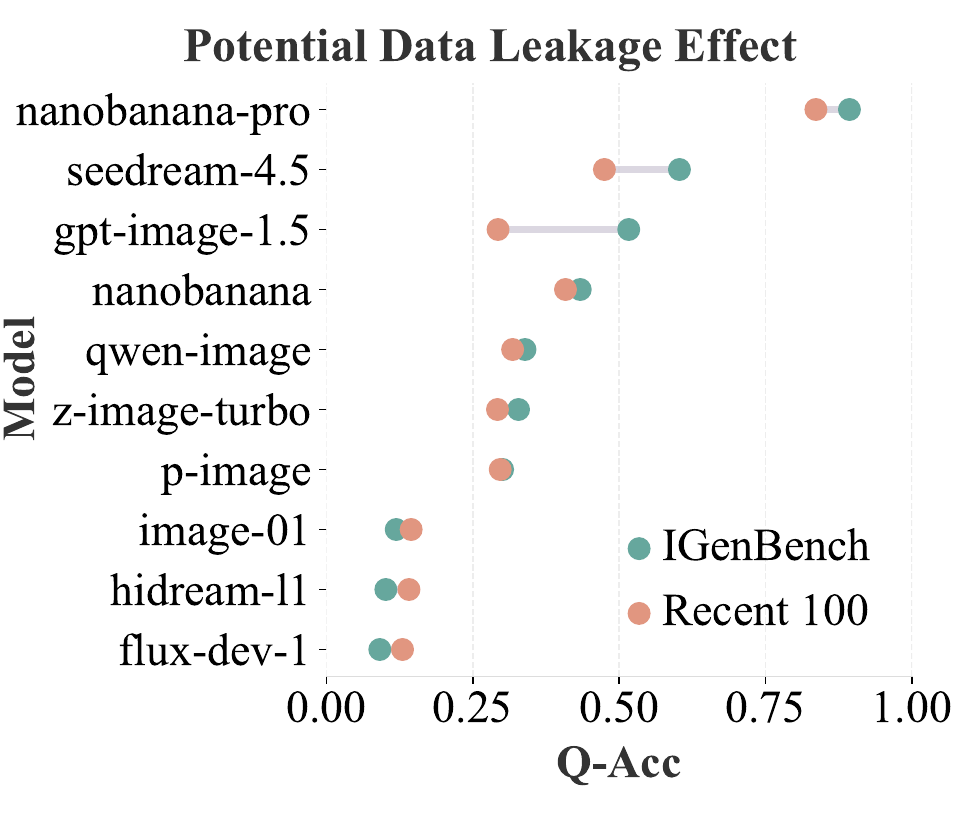}
    \caption{Potential data leakage effect. Q-ACC comparison between the original \system benchmark (green) and 100 recently published infographics from after December 2025 (orange).}
    \label{fig:Potential Data Leakage Effect}
\end{figure}

To assess the potential impact of data leakage on our benchmark evaluation, we conducted an additional experiment using 100 recently published infographics from Visual Capitalist dated after December 2025—ensuring that all samples were created after the release dates of the evaluated models. This temporal separation guarantees that these infographics could not have been included in any model's training data. Figure~\ref{fig:Potential Data Leakage Effect} presents the Q-ACC comparison between our original \system benchmark (green dots) and the recent 100 samples (orange dots). The results reveal two key findings: (1) Most models demonstrate stable performance, with the majority showing consistent Q-ACC scores across both datasets and an average change of only 0.7\%. This stability validates the reliability and generalizability of \system's evaluation framework, suggesting that our benchmark accurately reflects model capabilities rather than memorization artifacts. (2) GPT-Image-1.5 exhibits significant performance degradation, with Q-Acc dropping substantially from 0.52 to 0.29 on recent samples, indicating possible data contamination in the benchmark. We acknowledge that some degree of data leakage may exist for certain models, particularly those with more recent training cutoffs. To address this limitation, we plan to evolve \system into a live benchmark in future work.

\subsection{Error Analysis of Automated Evaluation}
\label{sec:error_analysis}

To better understand the reliability and limitations of our automated evaluation approach, we conduct a fine-grained error analysis of Gemini-2.5-Pro's judgments across different question categories. We manually examine all instances where the automated evaluator disagrees with human annotators, categorizing errors into two types: \textit{false-positive} (where the model incorrectly marks a violation as correct) and \textit{false-negative} (where the model incorrectly flags a correct element as wrong).

\begin{figure}[htbp]
    \centering
    \includegraphics[width=\columnwidth]{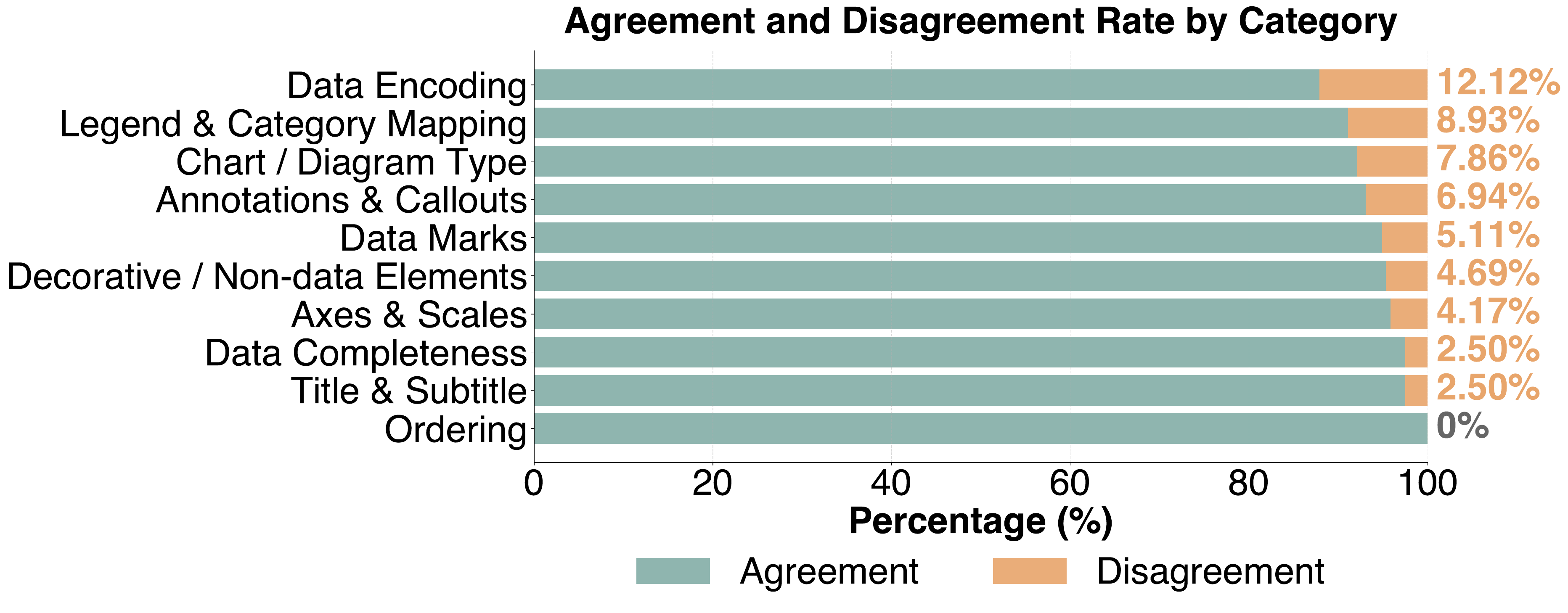}
    \caption{Agreement and disagreement rates between Gemini-2.5-Pro and human annotators across different question categories.}
    \label{fig:error_analysis}
\end{figure}

As shown in Figure~\ref{fig:error_analysis}, disagreement rates vary substantially across categories. Data Encoding exhibits the highest disagreement rate at 12.12\%, primarily due to over-positive errors where the evaluator fails to detect subtle encoding violations. In contrast, categories such as Title \& Subtitle (2.50\%), Data Completeness (2.50\%), and Ordering (0.00\%) demonstrate near-perfect agreement with human judgments. These results indicate that while Gemini-2.5-Pro achieves strong overall alignment with human evaluation, certain fine-grained data encoding aspects remain challenging for automated assessment.

\subsection{Evaluator Bias Investigation}
\label{sec:evaluator_bias}

Using a Google model (Gemini-2.5-Pro) as the primary evaluator could introduce bias toward Google generators. To investigate this, we selected the three evaluator MLLMs with the highest human alignment from our evaluator comparison study (Figure~\ref{fig:corr}): Gemini-2.5-Pro, GLM-4.5V, and GPT-5-Mini, spanning three different providers (Google, Zhipu AI, OpenAI). We randomly sampled one item per chart type (30 items, 2,780 evaluation questions) and evaluated all 10 generation models with each evaluator. As shown in Table~\ref{tab:evaluator_bias}, while absolute Q-ACC values differ across evaluators, model rankings are highly stable: all pairwise Spearman $\rho \geq 0.95$. Notably, Gemini-2.5-Pro does not favor Google generators, as it assigns the lowest absolute scores for Nanobanana-Pro and Nanobanana compared to other evaluators. The high ranking consistency across evaluators from three different providers suggests that our main results are not artifacts of evaluator-specific bias.

\begin{table}[h]
\centering
\small
\setlength{\tabcolsep}{4pt}
\renewcommand{\arraystretch}{1.1}
\caption{Q-ACC and model rankings across three evaluators from different providers. All pairwise Spearman $\rho \geq 0.95$.}
\label{tab:evaluator_bias}
\begin{tabular}{lccc}
\toprule
\textbf{Model} & \textbf{Gemini} & \textbf{GLM} & \textbf{GPT-5} \\
\midrule
Nanobanana-Pro & 0.89 (\#1) & 0.91 (\#1) & 0.92 (\#1) \\
Seedream-4.5   & 0.62 (\#2) & 0.73 (\#2) & 0.69 (\#2) \\
GPT-Image-1.5  & 0.47 (\#3) & 0.57 (\#4) & 0.52 (\#4) \\
Nanobanana     & 0.42 (\#4) & 0.63 (\#3) & 0.53 (\#3) \\
Z-Image-Turbo  & 0.36 (\#5) & 0.50 (\#6) & 0.44 (\#6) \\
P-Image        & 0.35 (\#6) & 0.49 (\#7) & 0.41 (\#7) \\
Qwen-Image     & 0.33 (\#7) & 0.51 (\#5) & 0.45 (\#5) \\
HIDream-I1     & 0.12 (\#8) & 0.19 (\#8) & 0.21 (\#8) \\
FLUX.1-dev     & 0.12 (\#9) & 0.14 (\#9) & 0.19 (\#9) \\
Image-01       & 0.11 (\#10) & 0.14 (\#10) & 0.13 (\#10) \\
\bottomrule
\end{tabular}
\end{table}

\subsection{Reference-Generation Visual Similarity}
\label{sec:ref_similarity}

To examine whether reference infographics carry evaluative signal, we compute four image similarity metrics between each generated infographic and its human reference across all 600 benchmark items and 10 models: CLIP Image Similarity, SSIM, PSNR, and LPIPS. As shown in Table~\ref{tab:ref_similarity_models}, these similarity metrics show limited discriminative power: most values fall within a narrow range (e.g., CLIP Similarity spans 0.67--0.80), and models with similar scores can differ drastically in Q-ACC (e.g., Z-Image-Turbo and Nanobanana-Pro both score 0.80 in CLIP Similarity but differ by over 55 points in Q-ACC). Different metrics also produce inconsistent model rankings. We further examine whether per-item similarity predicts per-item Q-ACC by computing Spearman rank correlation across all 7,096 (item, model) pairs, as shown in Table~\ref{tab:ref_similarity_corr}. Perceptual-level metrics (CLIP, LPIPS) show moderate positive correlations with Q-ACC, while pixel-level metrics (SSIM, PSNR) are near zero. These results confirm that while reference infographics carry some evaluative signal, our QA-based evaluation captures substantially more than visual resemblance, supporting the design choice of decomposed QA evaluation over reference-based scoring.

\begin{table}[h]
\centering
\small
\setlength{\tabcolsep}{3.5pt}
\caption{Model-level average similarity between generated infographics and human references. Rankings are shown in parentheses.}
\label{tab:ref_similarity_models}
\begin{tabular}{lcccc}
\toprule
\textbf{Model} & \textbf{CLIP} & \textbf{SSIM} & \textbf{PSNR} & \textbf{LPIPS}$\downarrow$ \\
\midrule
Nanobanana      & 0.80 (\#1) & 0.52 (\#1) & 8.45 (\#2) & 0.64 (\#3) \\
Nanobanana-Pro  & 0.80 (\#1) & 0.48 (\#5) & 8.45 (\#2) & 0.61 (\#1) \\
Z-Image-Turbo   & 0.80 (\#1) & 0.50 (\#3) & 7.73 (\#5) & 0.64 (\#3) \\
Seedream-4.5    & 0.78 (\#4) & 0.46 (\#6) & 7.61 (\#6) & 0.66 (\#5) \\
Qwen-Image      & 0.78 (\#4) & 0.44 (\#8) & 6.93 (\#8) & 0.67 (\#6) \\
P-Image         & 0.77 (\#6) & 0.46 (\#6) & 7.16 (\#7) & 0.68 (\#7) \\
GPT-Image-1.5   & 0.77 (\#6) & 0.39 (\#10) & 7.07 (\#8) & 0.67 (\#6) \\
HIDream-I1      & 0.75 (\#8) & 0.41 (\#9) & 5.96 (\#10) & 0.70 (\#10) \\
FLUX.1-dev      & 0.70 (\#9) & 0.49 (\#4) & 8.64 (\#1) & 0.68 (\#7) \\
Image-01        & 0.67 (\#10) & 0.52 (\#1) & 8.58 (\#4) & 0.69 (\#9) \\
\bottomrule
\end{tabular}
\end{table}

\begin{table}[h]
\centering
\small
\caption{Spearman correlation between per-item image similarity metrics and Q-ACC across 7,096 (item, model) pairs.}
\label{tab:ref_similarity_corr}
\begin{tabular}{lcc}
\toprule
\textbf{Similarity Metric} & \textbf{Spearman $\rho$} & \textbf{$p$-value} \\
\midrule
CLIP Image Sim. & 0.30 & $<10^{-6}$ \\
LPIPS ($\downarrow$=better) & $-$0.22 & $<10^{-6}$ \\
PSNR & 0.08 & $<10^{-6}$ \\
SSIM & 0.05 & $<10^{-6}$ \\
\bottomrule
\end{tabular}
\end{table}

\subsection{Prompt Length Sensitivity}
\label{sec:prompt_length}

We analyze how prompt complexity affects model performance across all 600 benchmark items and 10 models. We measure three complexity variables for each prompt: total token count (\textit{total\_length}), token count of the embedded data portion (\textit{data\_length}), and token count of the layout/visual specification (\textit{semantic\_length}). As shown in Table~\ref{tab:prompt_length}, all three variables show significant negative correlations with Q-ACC, with \textit{total\_length} and \textit{data\_length} exhibiting the strongest effects ($\rho \approx -0.57$). This confirms that data volume dominates prompt length and is the primary source of difficulty. Models also differ substantially in their sensitivity: GPT-Image-1.5 is most sensitive ($\rho = -0.550$), while Nanobanana-Pro is most robust ($\rho = -0.198$).

\begin{table}[h]
\centering
\small
\caption{Spearman correlation between prompt complexity variables and mean Q-ACC across 600 items.}
\label{tab:prompt_length}
\begin{tabular}{lcc}
\toprule
\textbf{Variable} & \textbf{Spearman $\rho$} & \textbf{$p$-value} \\
\midrule
total\_length & $-$0.573 & $1.4 \times 10^{-53}$ \\
data\_length & $-$0.557 & $4.3 \times 10^{-50}$ \\
semantic\_length & $-$0.403 & $7.1 \times 10^{-25}$ \\
\bottomrule
\end{tabular}
\end{table}

\subsection{Question Type Independence}
\label{sec:question_independence}

To verify that the 10 question types are not redundant, we compute within-model pairwise Spearman correlations on per-item accuracy across all 600 items, then average across all 10 models. Computing correlations within each model avoids inflated correlations caused by pooling models of different overall ability. As shown in Table~\ref{tab:question_independence}, among all dimension pairs, the highest within-model correlation is only 0.21 (Data Completeness vs.\ Data Encoding, and Data Completeness vs.\ Data Marks). Most pairs fall below 0.10, and many are near zero (e.g., Decorative vs.\ Completeness: 0.01, Ordering vs.\ Title: 0.02). ``--'' indicates cases where some weak models have near-zero variance on that dimension, preventing correlation computation. These results confirm that the 10 question types capture largely independent aspects of infographic fidelity with minimal redundancy.

\begin{table}[h]
\centering
\scriptsize
\setlength{\tabcolsep}{2.5pt}
\caption{Average within-model pairwise Spearman correlation matrix across the 10 question types. All correlations remain below 0.21. ``--'' indicates near-zero variance preventing computation.}
\label{tab:question_independence}
\begin{tabular}{lcccccccccc}
\toprule
 & \rotatebox{70}{\textbf{Comp.}} & \rotatebox{70}{\textbf{Enc.}} & \rotatebox{70}{\textbf{Order}} & \rotatebox{70}{\textbf{Marks}} & \rotatebox{70}{\textbf{Anno.}} & \rotatebox{70}{\textbf{Axes}} & \rotatebox{70}{\textbf{Leg.}} & \rotatebox{70}{\textbf{Chart}} & \rotatebox{70}{\textbf{Title}} & \rotatebox{70}{\textbf{Deco.}} \\
\midrule
Comp.  & 1.00 & .21 & --  & .21 & .06 & --  & --  & .15 & .04 & .01 \\
Enc.   & .21 & 1.00 & .20 & .16 & .08 & .01 & .05 & .10 & .03 & .02 \\
Order  & --  & .20 & 1.00 & --  & .09 & --  & --  & .18 & .02 & --  \\
Marks  & .21 & .16 & --  & 1.00 & .09 & .08 & .06 & .10 & .04 & .04 \\
Anno.  & .06 & .08 & .09 & .09 & 1.00 & .05 & .06 & .04 & .11 & .07 \\
Axes   & --  & .01 & --  & .08 & .05 & 1.00 & --  & .05 & .09 & .04 \\
Leg.   & --  & .05 & --  & .06 & .06 & --  & 1.00 & .07 & .03 & .04 \\
Chart  & .15 & .10 & .18 & .10 & .04 & .05 & .07 & 1.00 & .04 & .03 \\
Title  & .04 & .03 & .02 & .04 & .11 & .09 & .03 & .04 & 1.00 & .01 \\
Deco.  & .01 & .02 & --  & .04 & .07 & .04 & .04 & .03 & .01 & 1.00 \\
\bottomrule
\end{tabular}
\end{table}

\section{Representative Cases}
\label{appendix:full_case}

\begin{figure*}[t]
  \centering
  \includegraphics[width=\linewidth]{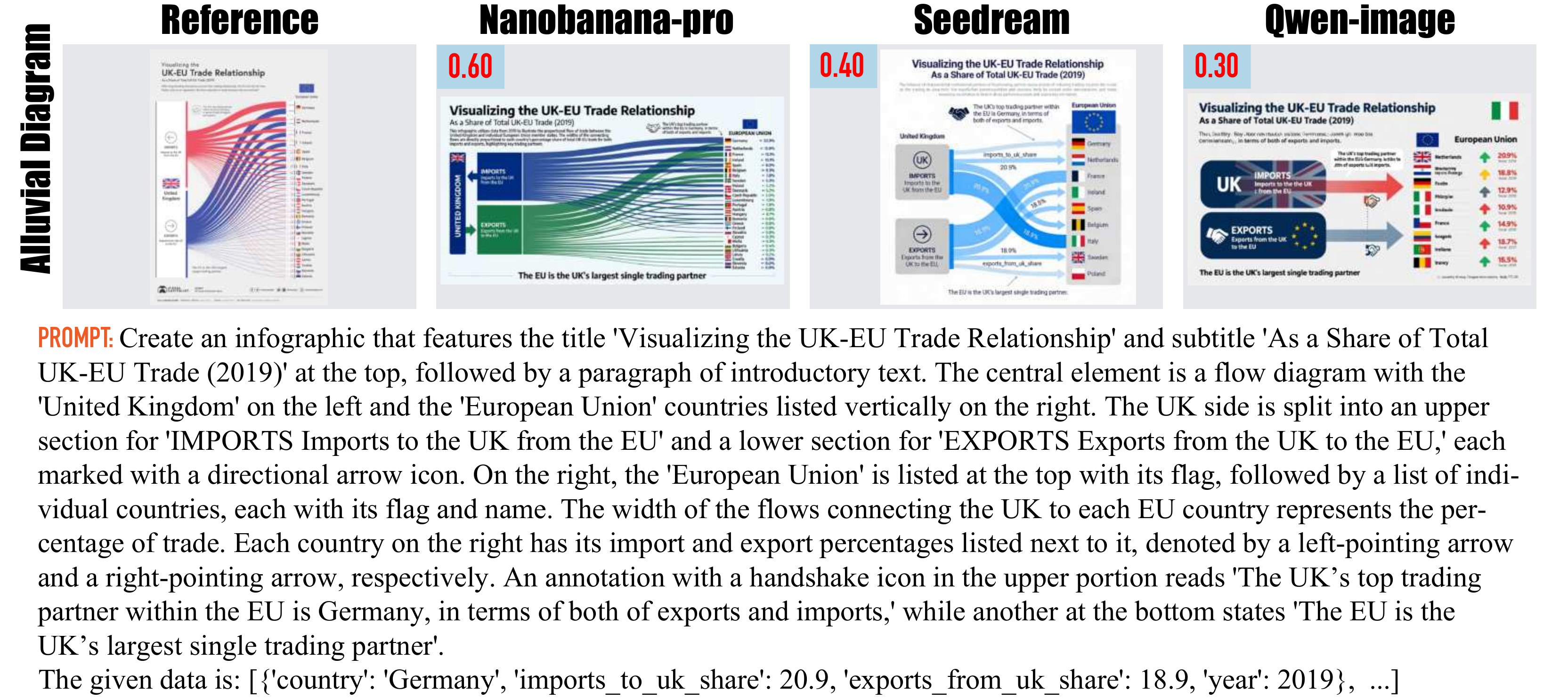}
  \caption{Case of Alluvial Diagram.}
  \label{fig:case_alluvial_diagram}
\end{figure*}

\begin{figure*}[t]
  \centering
  \includegraphics[width=\linewidth]{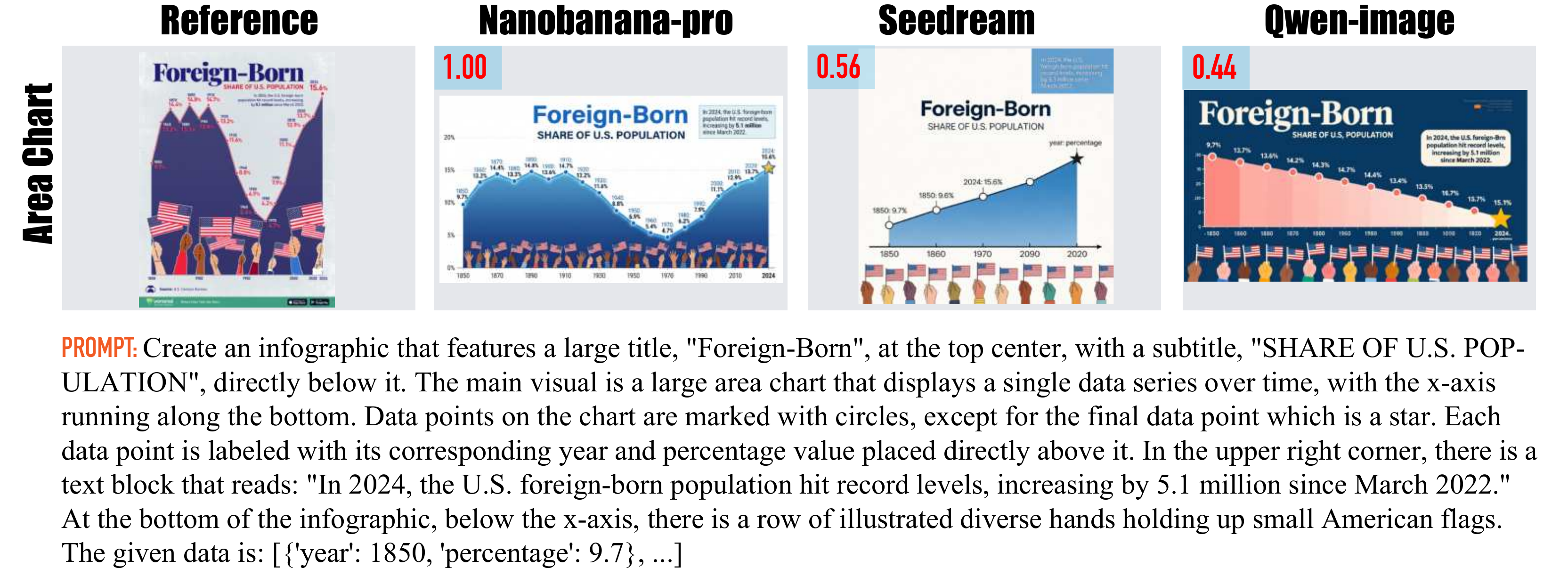}
  \caption{Case of Area Chart.}
  \label{fig:case_area_chart}
\end{figure*}

\begin{figure*}[t]
  \centering
  \includegraphics[width=\linewidth]{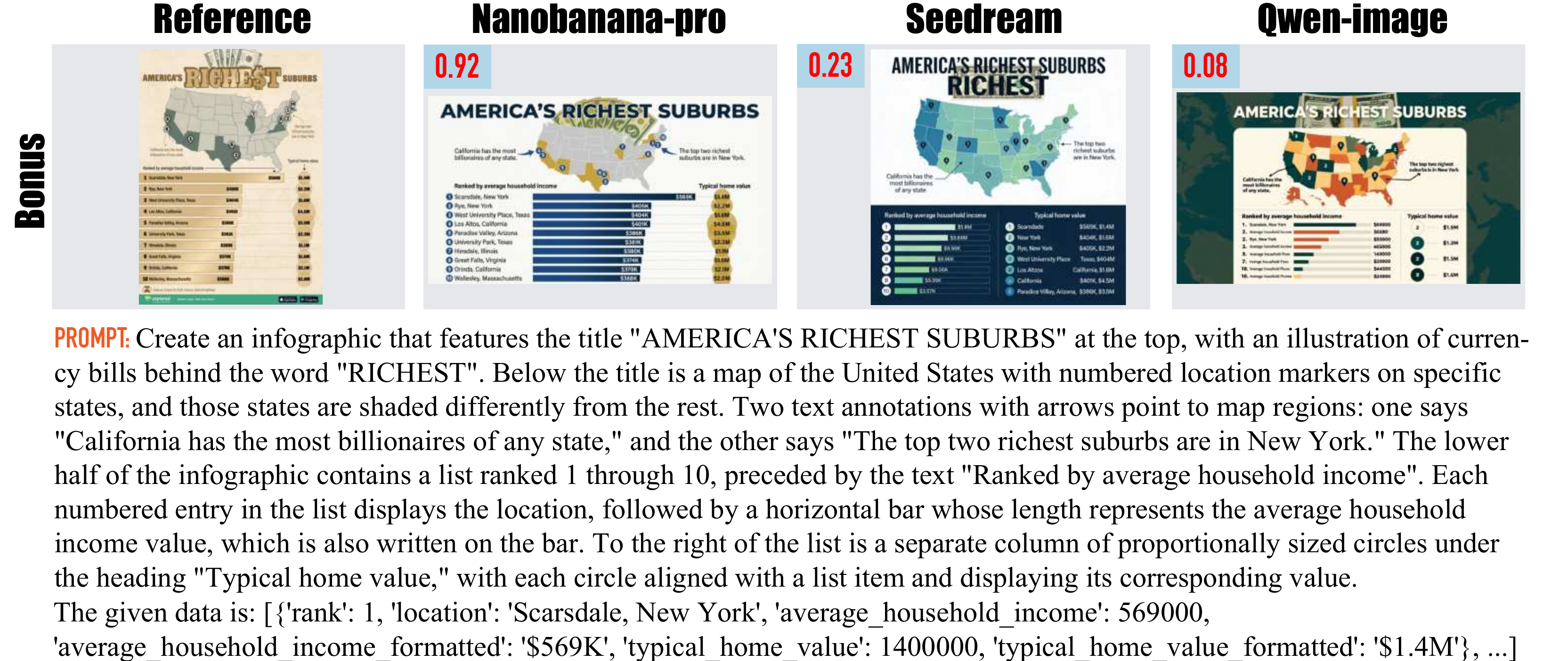}
  \caption{Case of Bonus.}
  \label{fig:case_bonus}
\end{figure*}

\begin{figure*}[t]
  \centering
  \includegraphics[width=\linewidth]{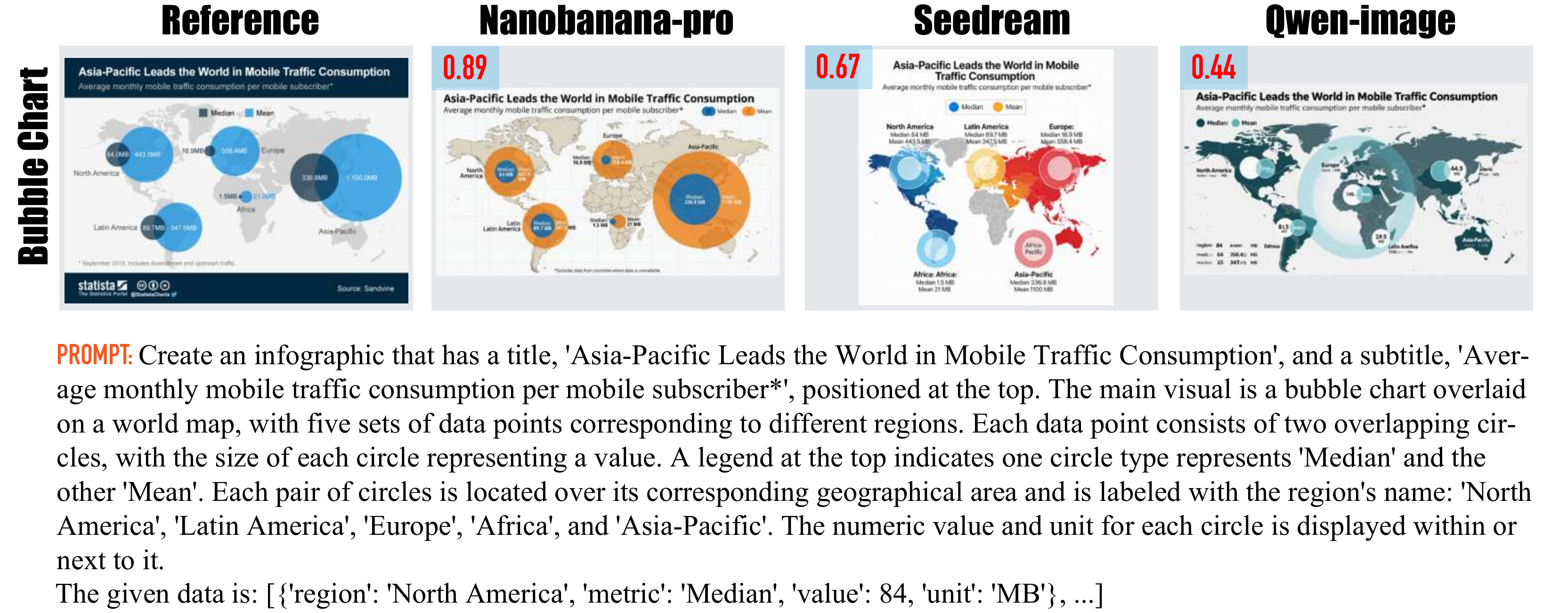}
  \caption{Case of Bubble Chart.}
  \label{fig:case_bubble_chart}
\end{figure*}

We present four additional representative case studies in Figures~\ref{fig:case_alluvial_diagram} to~\ref{fig:case_bubble_chart}.
Together with the proportional area chart case in Figure~\ref{fig:case_study_proportional_area_chart}, these examples illustrate the visual diversity of \system without repeating the full gallery in the paper.
The complete set of case studies covering all 30 chart types is available on the project website at \url{https://igen-bench.vercel.app/}.

\section{Instruction and Prompts}
\label{appendix:prompts}
\begin{figure*}[t]
  \centering
  \includegraphics[width=\textwidth]{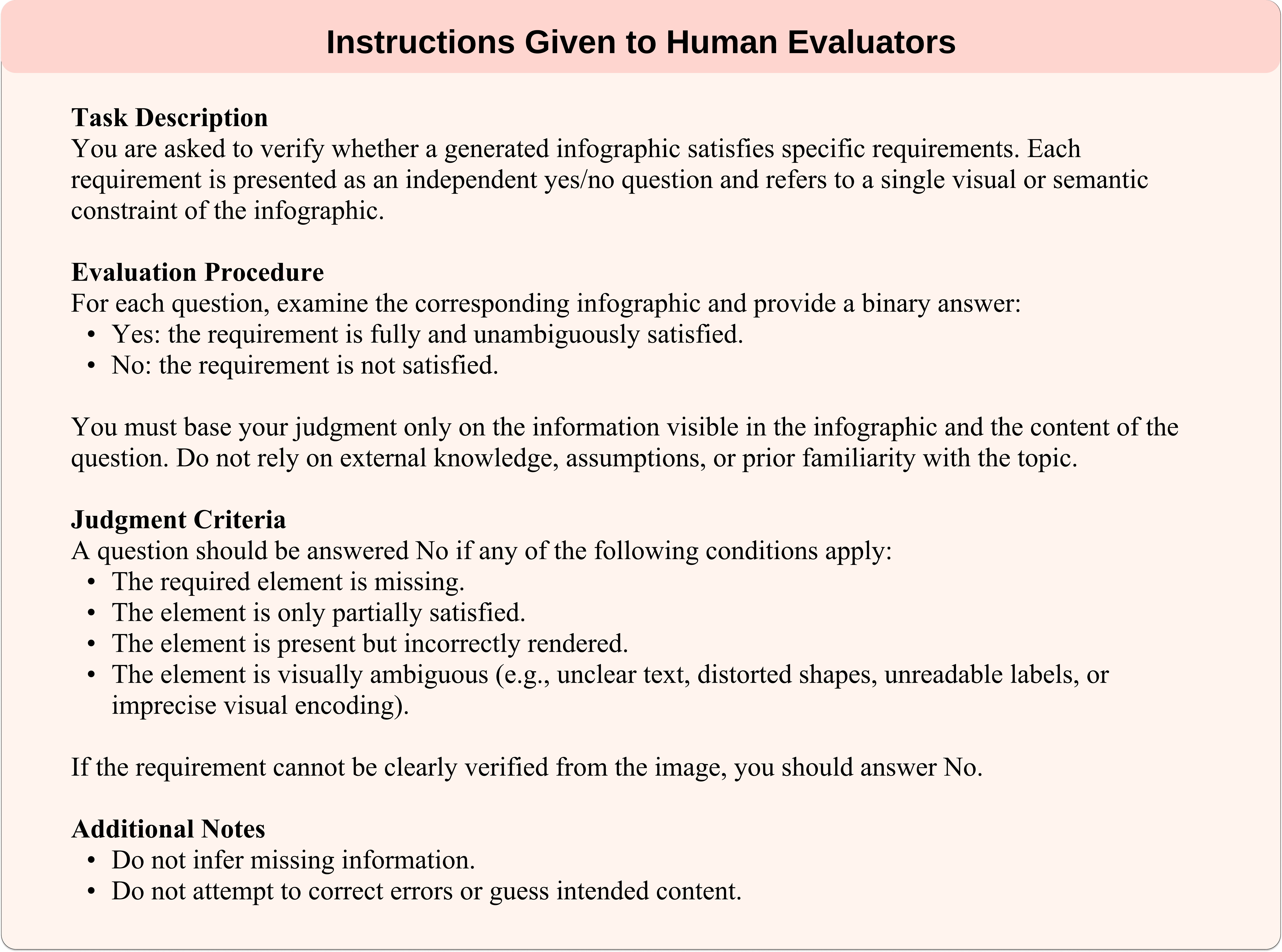}
  \caption{Instructions provided to human evaluators for the annotation task.}
  \label{fig:instruction_human}
\end{figure*}

\begin{figure*}[t]
  \centering
  \includegraphics[width=\linewidth]{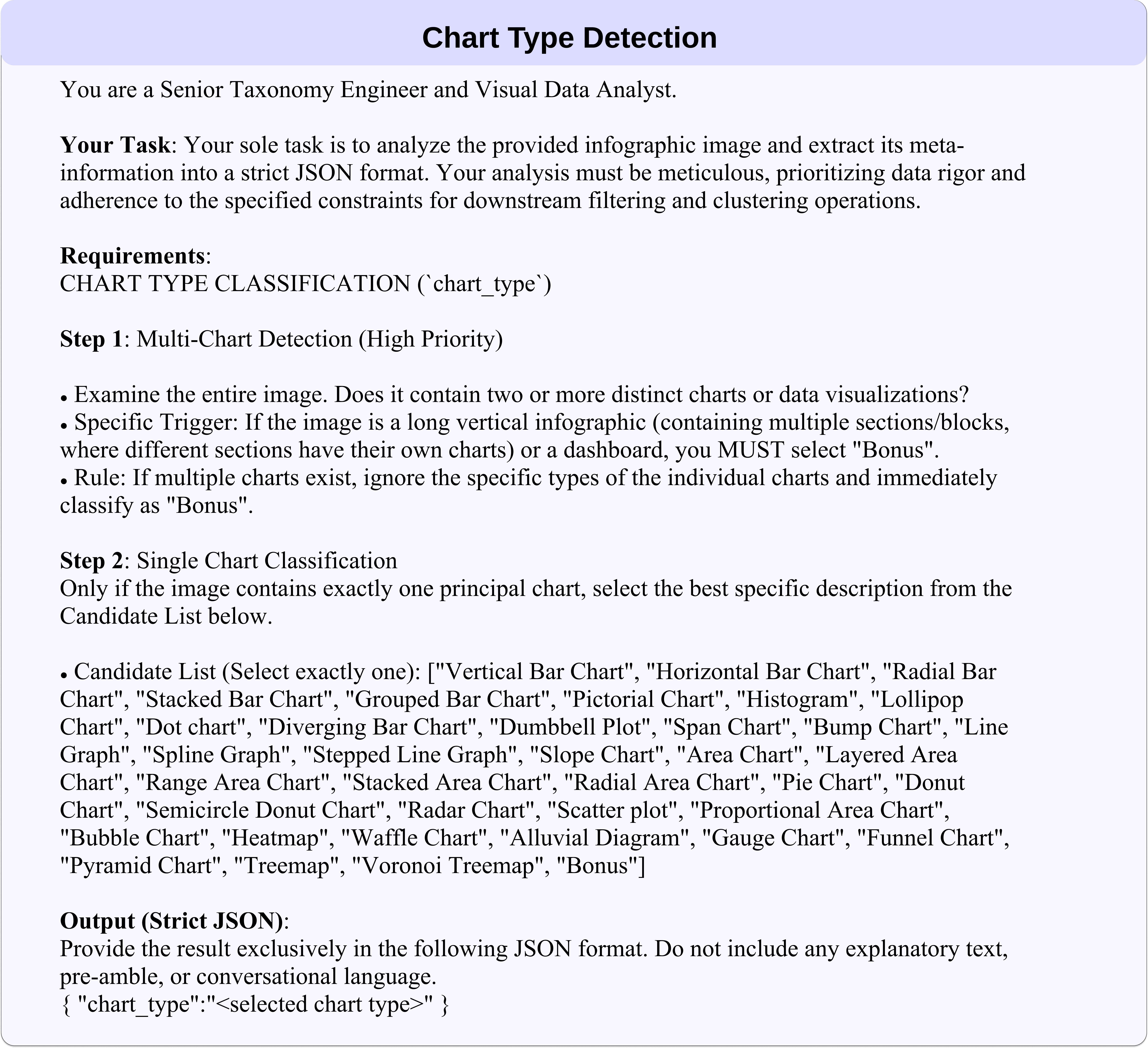}
  \caption{Prompt for chart type detection.}
  \label{fig:chart_type_detection}
\end{figure*}

\begin{figure*}[t]
  \centering
  \includegraphics[width=\linewidth]{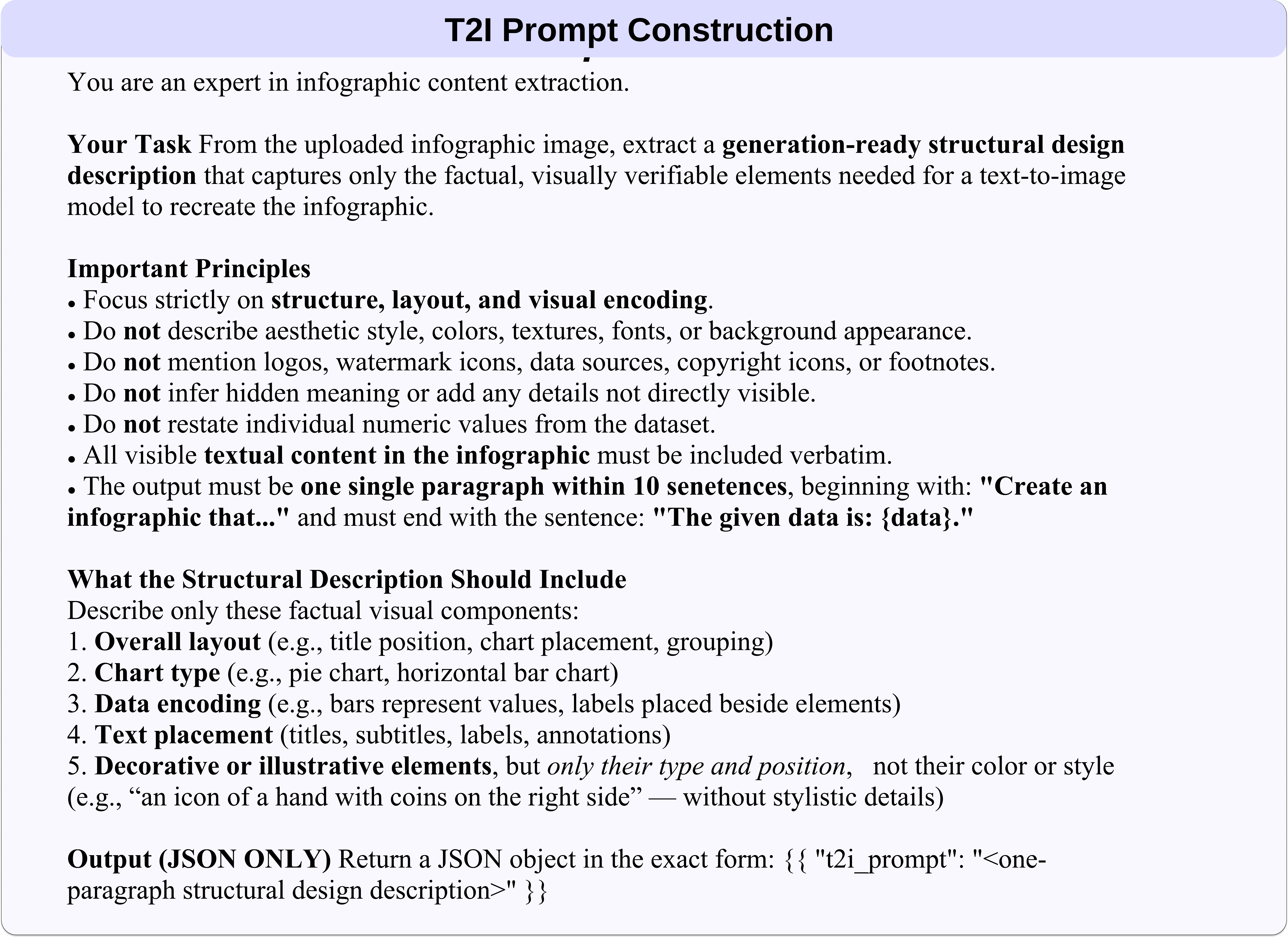}
  \caption{Prompt used in T2I prompt construction.}
  \label{fig:t2i_construction}
\end{figure*}

\begin{figure*}[t]
  \centering
  \includegraphics[width=\linewidth]{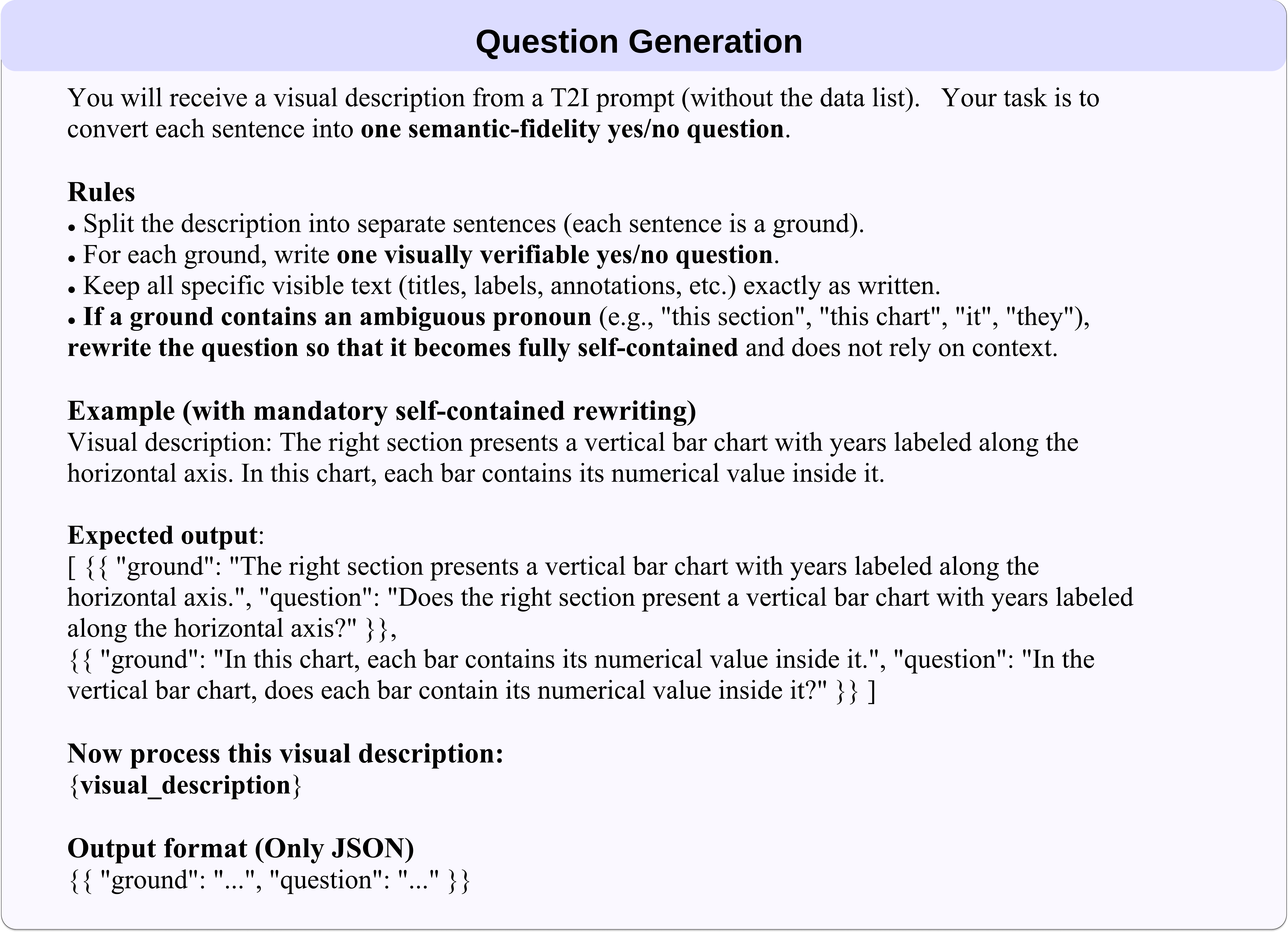}
  \caption{Prompt for question generation.}
  \label{fig:question_generation}
\end{figure*}

\begin{figure*}[t]
  \centering
  \includegraphics[width=\linewidth]{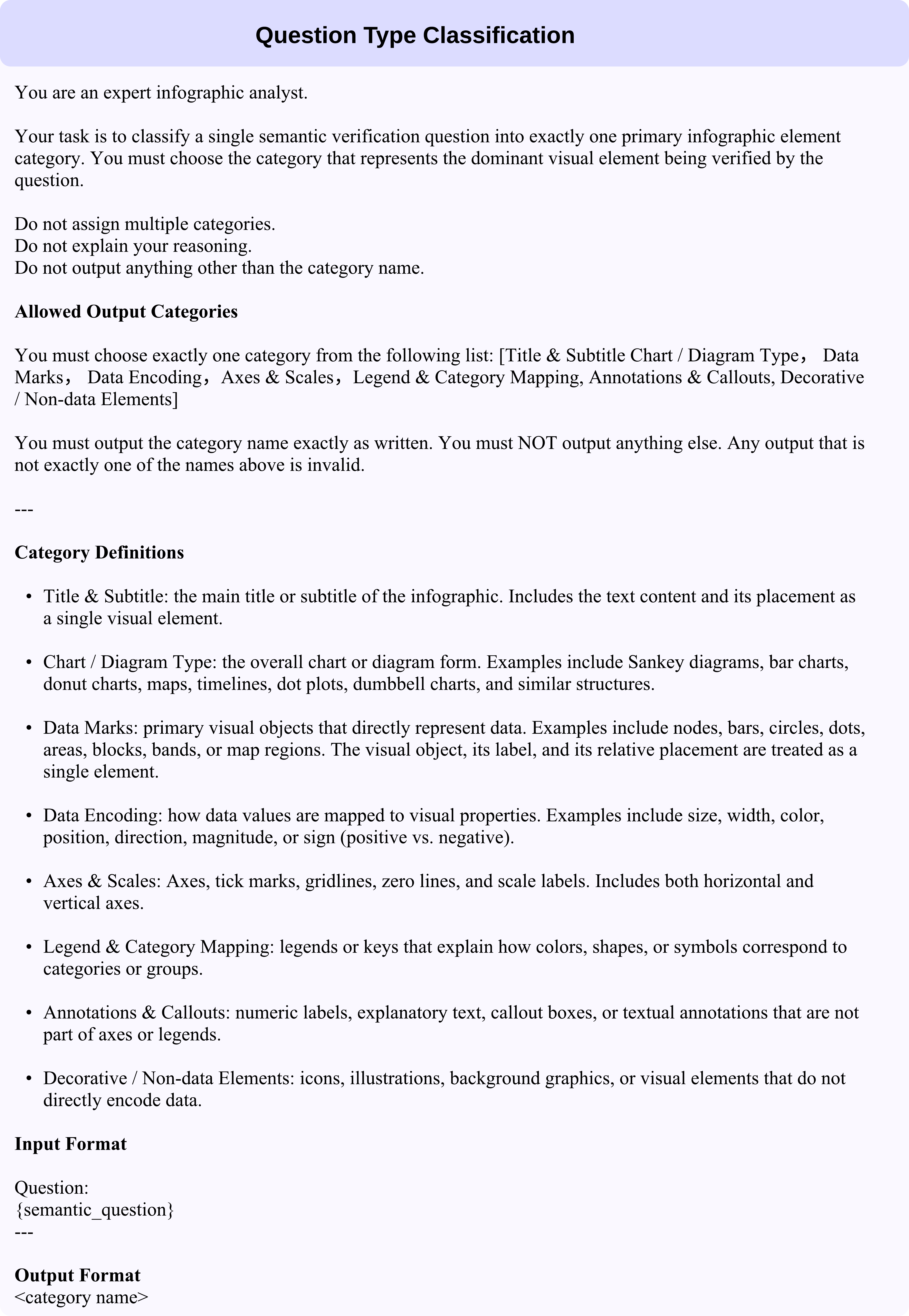}
  \caption{Prompt for question type classification.}
  \label{fig:question_type_classification}
\end{figure*}

\begin{figure*}[t]
  \centering
  \includegraphics[width=\linewidth]{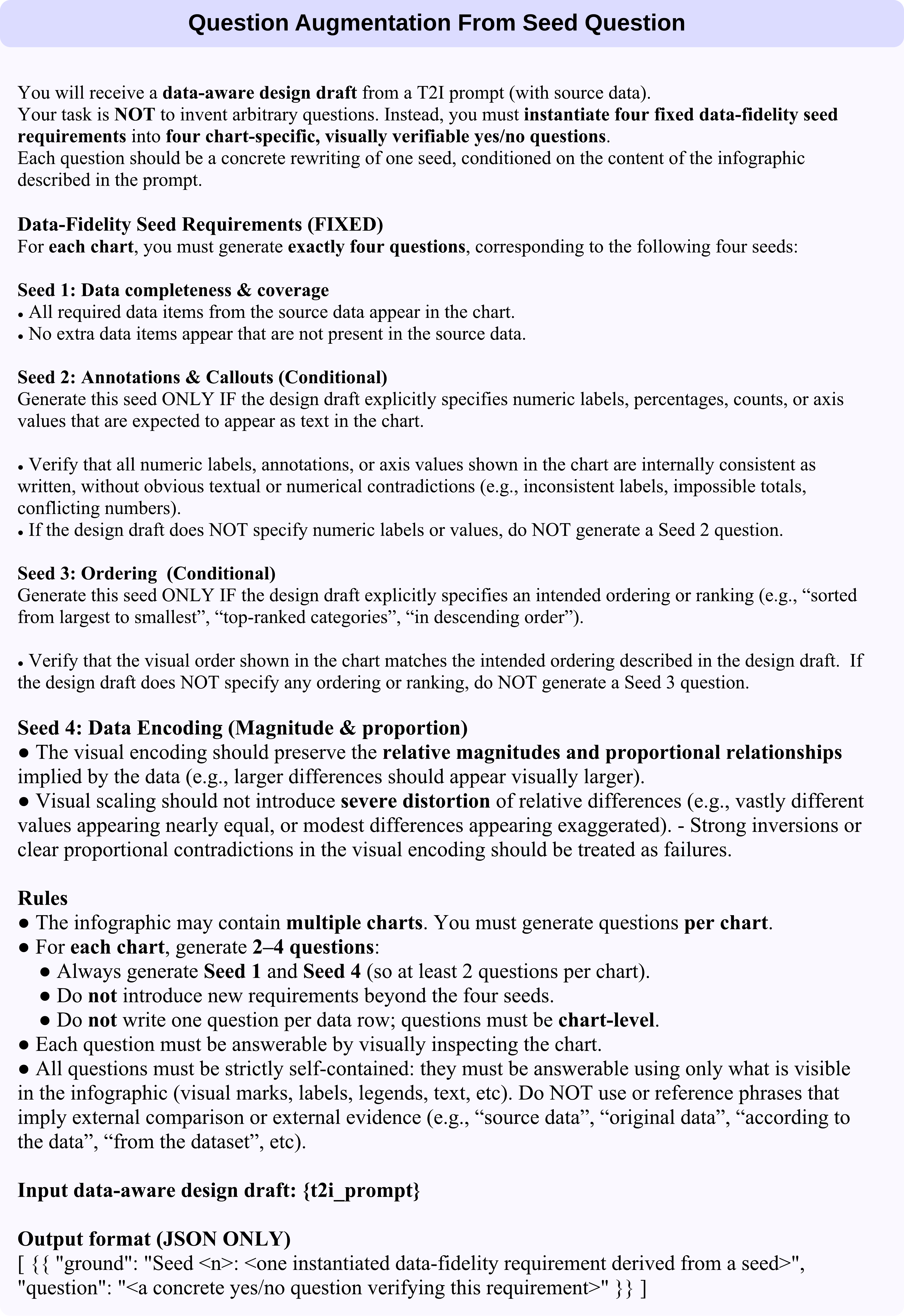}
  \caption{Prompt used for question augmentation from seed questions.}
  \label{fig:question_augmentation}
\end{figure*}

\begin{figure*}[t]
  \centering
  \includegraphics[width=\linewidth]{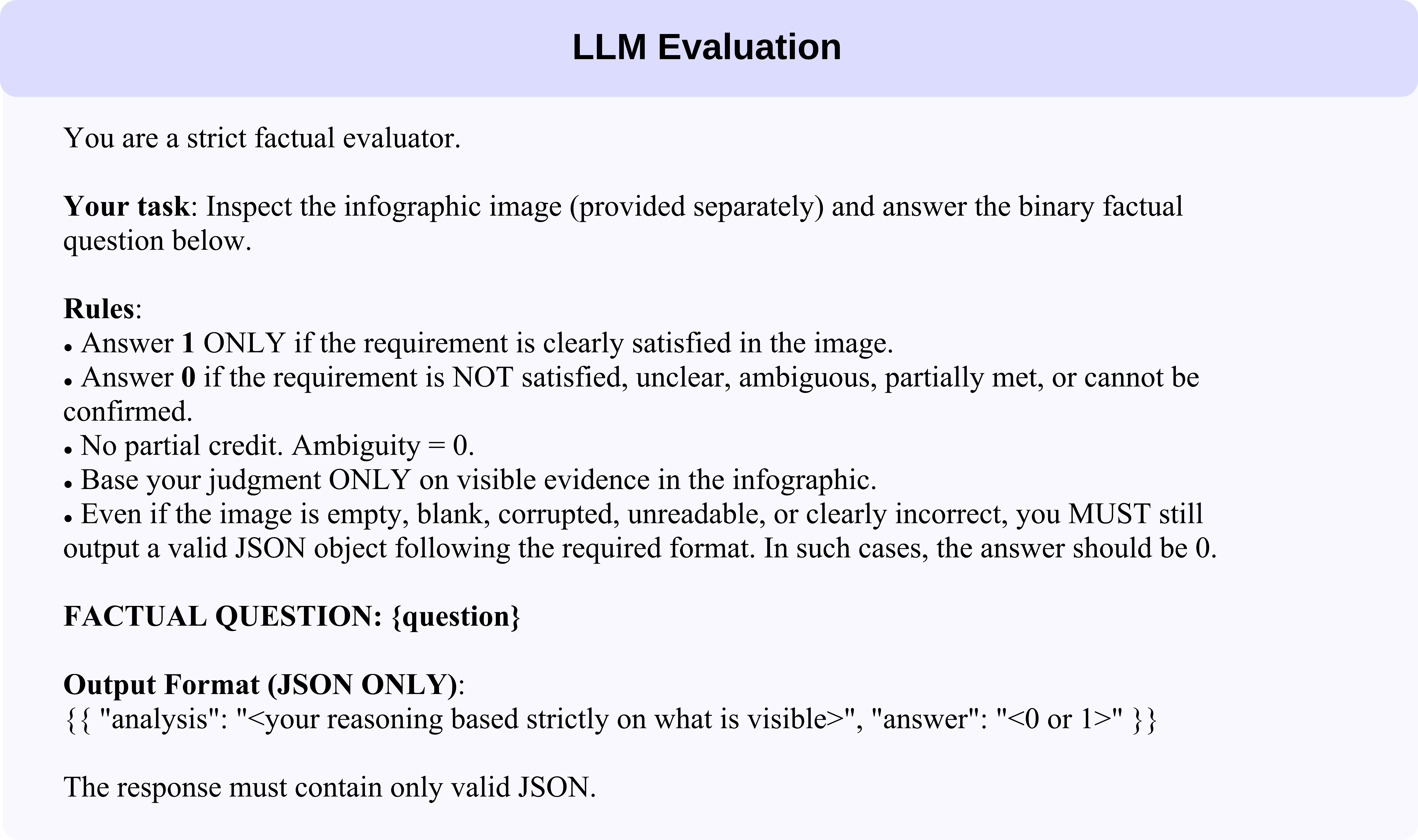}
  \caption{Prompt used in LLM evaluation.}
  \label{fig:llm_evaluation}
\end{figure*}

In this section, we provide the instruction given to human evaluators, along with all prompts used throughout our pipeline for transparency and reproducibility. Figure~\ref{fig:instruction_human} shows the task instruction provided to human evaluators during the evaluation phase. Figure~\ref{fig:chart_type_detection} presents the prompt for chart type detection. Figure~\ref{fig:t2i_construction} shows the prompt for T2I prompt construction. Figure~\ref{fig:question_generation} illustrates the prompt for question generation, decomposing T2I prompts into atomic verification questions. Figure~\ref{fig:question_type_classification} displays the prompt for question type classification, categorizing generated questions into our predefined taxonomy. Figure~\ref{fig:question_augmentation} presents the prompt for question augmentation, which expands the question set based on seed questions to ensure comprehensive coverage. Finally, Figure~\ref{fig:llm_evaluation} shows the prompt used for automated evaluation of generated infographics against verification questions.

\clearpage

\end{document}